\documentclass[journal]{IEEEtran}

\usepackage{graphicx}
\usepackage{booktabs}
\usepackage{multirow}
\usepackage{array}
\usepackage{tabularx}
\usepackage{makecell}
\usepackage{cite}
\usepackage{amsmath,amssymb,amsfonts}
\usepackage{amsthm}
\usepackage{mathrsfs}
\usepackage[table,xcdraw]{xcolor}
\usepackage{textcomp}
\usepackage{colortbl}
\usepackage{hhline}
\usepackage{adjustbox}
\usepackage{wrapfig}
\usepackage{arydshln}
\usepackage{threeparttable}
\usepackage[utf8]{inputenc}
\usepackage[T1]{fontenc}
\usepackage{url}
\usepackage{nicefrac}
\usepackage{microtype}
\usepackage{stfloats}
\usepackage{balance}
\usepackage{pifont}
\usepackage[hidelinks]{hyperref}
\hypersetup{
    pdftitle={Gaze-DETR: Top-Down Guidance Through Priority Maps for Infrared Weak-Small UAV Detection with DETR},
    pdfauthor={Nian Liu et al.},
    pdfkeywords={thermal infrared UAV detection, weak-small target detection, eye tracking, top-down guidance, spatial priority map}
}
\usepackage{placeins}

\newcommand{\cmark}{\textcolor{teal!80!black}{\ding{51}}}

\newcommand{\abyes}{\ensuremath{\checkmark}}
\newcommand{\abno}{\ensuremath{\times}}
\newcommand{\PaperTableFont}{\scriptsize}
\newcommand{\abgain}[1]{\textcolor{green!50!black}{\scriptsize$\uparrow$#1}}
\newcommand{\abdrop}[1]{\textcolor{red!70!black}{\scriptsize$\downarrow$#1}}
\newcommand{\attrgain}[1]{\textcolor{green!50!black}{\tiny$\uparrow$#1}}
\newcommand{\attrloss}[1]{\textcolor{red!70!black}{\tiny$\downarrow$#1}}
\newcommand{\attrzero}{\textcolor{black!60}{\tiny$0.00$}}

\definecolor{darkgreen}{rgb}{0.0, 0.5, 0.0}
\begin{document}

\title{Gaze-DETR: Top-Down Guidance Through Priority Maps for Infrared Weak-Small UAV \\Detection with DETR}

% \author{Nian~Liu$^{*}$, Yuxin~Yang, Shubo~Lin, Sikui~Zhang, Boyu~Cai, Liang~Li, Yizheng~Wang, Weiming~Hu,~\IEEEmembership{Senior Member,~IEEE}, and Jin~Gao$^{\dagger}$%
% \thanks{N. Liu and Y. Yang contributed equally to this work. Corresponding author: Jin Gao (jin.gao@nlpr.ia.ac.cn).}%
% \thanks{N. Liu is with the School of Advanced Interdisciplinary Sciences, University of Chinese Academy of Sciences, Beijing 101408, China, and also with the State Key Laboratory of Multimodal Artificial Intelligence Systems, Institute of Automation, Chinese Academy of Sciences, Beijing 100190, China.}%
% \thanks{J. Gao, Y. Yang, S. Zhang, and W. Hu are with the State Key Laboratory of Multimodal Artificial Intelligence Systems, Institute of Automation, Chinese Academy of Sciences, Beijing 100190, China, and also with the School of Artificial Intelligence, University of Chinese Academy of Sciences, Beijing 101408, China. B. Cai and W. Hu are also with the School of Information Science and Technology, ShanghaiTech University, Shanghai 201210, China.}%
% \thanks{L. Li and Y. Wang are with the Beijing Institute of Basic Medical Sciences, Beijing 100850, China.}%
% }
\author{
Nian~Liu,
Yuxin~Yang,
Shubo~Lin,
Sikui~Zhang,
Liang~Li,
Boyu~Cai,
% Liang~Li,
Yizheng~Wang,
Weiming~Hu,~\IEEEmembership{Senior Member,~IEEE},
and Jin~Gao%
\thanks{N. Liu, Y. Yang, S. Lin, S. Zhang, W. Hu, and J. Gao
are with the State Key Laboratory of Multimodal Artificial
Intelligence Systems, Institute of Automation, Chinese Academy
of Sciences, Beijing 100190, China. N. Liu is also with the
School of Advanced Interdisciplinary Sciences, University of
Chinese Academy of Sciences, Beijing 101408, China. S. Zhang
is also with the School of Artificial Intelligence, University
of Chinese Academy of Sciences, Beijing 101408, China. W. Hu
is also with the Beijing Key Laboratory of Super Intelligent
Security of Multi-Modal Information, Institute of Automation,
Chinese Academy of Sciences, Beijing 100190, China. B. Cai and
W. Hu are also with the School of Information Science and
Technology, ShanghaiTech University, Shanghai 201210, China.
Corresponding author: Jin Gao
(e-mail: jin.gao@nlpr.ia.ac.cn).}%
\thanks{L. Li and Y. Wang are with the Beijing Institute of Basic
Medical Sciences, Beijing 100850, China.}%
}

\IEEEaftertitletext{%
\vspace{-3.6em}
\begin{center}
\includegraphics[
    width=0.98\textwidth,
    trim=-1.5cm 0cm -2.5cm 0cm,
    clip
]{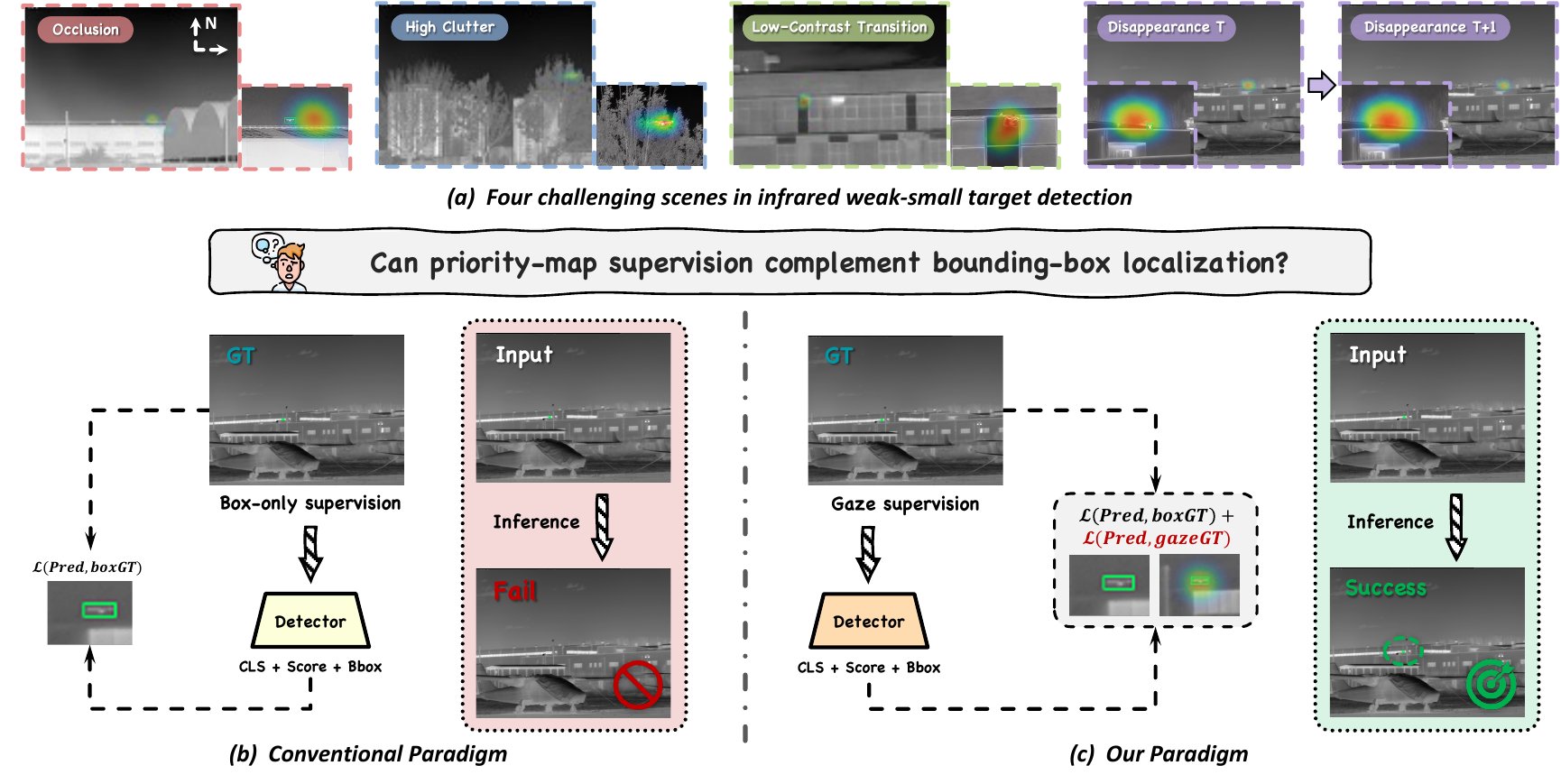}
\vspace{-1.5mm}
\refstepcounter{figure}\label{fig:motivation}%
\begin{minipage}{0.98\textwidth}
\footnotesize \textbf{Fig.~\thefigure.} Motivation for the proposed priority-supervision formulation. 
(a) Infrared weak-small UAV detection faces occlusion, clutter, low contrast, and apparent disappearance. 
(b) Box-only supervision provides final localization labels but limited pre-localization priority guidance. 
(c) Priority supervision complements box supervision by adding a training-time priority loss; the real-gaze scheme is shown as an example. 
At inference, only the infrared image is used. Green boxes denote ground-truth boxes, and heat maps denote fixation-density regions.
\end{minipage}
\vspace{3mm}
\end{center}
\vspace{-0.8em}
}

\markboth{IEEE Transactions on Geoscience and Remote Sensing}{Liu \MakeLowercase{\textit{et al.}}: Gaze-DETR: Top-Down Guidance Through Priority Maps for Infrared Weak-Small UAV Detection with DETR}

\maketitle

\vspace{-3mm}

\begin{abstract}
Infrared small target detection (ISTD) remains challenging because tiny, low-contrast targets are easily overwhelmed by clutter, noise, or occlusion.
Conventional single-frame and multi-frame detectors rely on bounding-box supervision, which specifies final locations but provides little explicit guidance for candidate prioritization or weak-target evidence preservation before localization.
Task-driven visual search offers such guidance: top-down goals and visual evidence jointly form a spatial priority map that ranks candidate locations.
Building on this principle, we propose Gaze-DETR, a bio-inspired detector that learns an internal priority map before localization.
First, a priority head predicts a normalized map from image features.
Second, Residual Priority-Guided Feature Modulation (RPFM) modulates high-priority responses while retaining multi-scale features.
Finally, Priority-Guided Anchor Query Injection (PAQI) converts high-priority locations into decoder anchor queries.
We train the priority head using three schemes: box-derived Gaussian priority-supervision maps; real-gaze priority-supervision maps constructed from fixation-density maps; and transferred pseudo-gaze priority-supervision maps obtained by learning gaze--box relations from paired annotations and applying them to Anti-UAV410 training boxes.
To support the latter two schemes, we construct TIR-UAV120-Gaze with paired detection and task-driven eye-tracking annotations.
On TIR-UAV120-Gaze, Gaze-DETR achieves 85.76 mAP$_{50}$ and 88.77 F1 with box-derived supervision, and 86.18 mAP$_{50}$ and 89.00 F1 with real-gaze supervision.
On Anti-UAV410, it achieves 87.06 mAP$_{50}$ and 90.90 F1 with box-derived supervision, and 87.08 mAP$_{50}$ and 90.43 F1 with transferred pseudo-gaze supervision.
These results show that explicit spatial-priority learning provides pre-localization guidance complementary to bounding-box supervision across annotation conditions and costs.
\end{abstract}


\begin{figure}[t]
    \centering
    \includegraphics[width=0.95\linewidth]{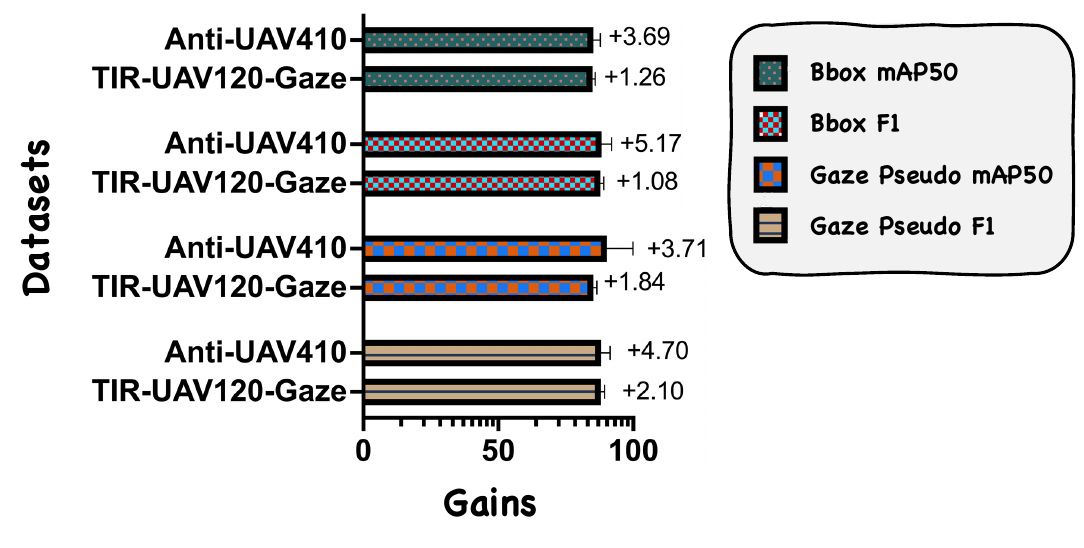}
    \caption{Effectiveness of the evaluated priority-supervision settings. Box-derived, real-gaze, and transferred pseudo-gaze supervision improve mAP$_{50}$ and F1 over DINO-DETR on their corresponding benchmarks while preserving image-only inference.}
    \label{fig:performance_gain_over_dino}
\end{figure}

\section{Introduction}
\label{sec:introduction}

\IEEEPARstart{I}{nfrared} small target detection (ISTD) aims to identify and localize targets that occupy only a few pixels in complex infrared scenes~\cite{li2023dnanet,zhang2022isnet}.
Within this field, detecting weak-small UAVs in thermal infrared imagery is important for wide-area surveillance, day-and-night monitoring, and anti-UAV warning~\cite{Wu_2020_CVPR_Workshops,jiang2021antiuav,huang2024antiuav410,zhu2023evidential}.
However, distant UAVs usually occupy only a very small image area and contain limited shape and texture information.
Moreover, their low target--background contrast and low signal-to-noise ratio produce extremely weak visual responses.
Consequently, these responses are easily overwhelmed by noise and clutter in complex infrared backgrounds.
Occlusion and apparent disappearance can further reduce target observability, as illustrated in Fig.~\ref{fig:motivation}(a).
Therefore, reliable weak-small UAV detection remains a challenging problem in practical infrared surveillance systems.

Existing ISTD methods can be broadly divided into model-driven and data-driven approaches.
Model-driven methods describe target--background differences using handcrafted operators or predefined structural priors.
Representative approaches include top-hat and max-mean filtering, local contrast methods, and low-rank and sparse models~\cite{Tophat,MaxMeandeshpande1999max,chen2013lcm,gao2013infrared,dai2017reweighted,Liu2020STT}.
These methods are generally interpretable and do not require large-scale annotated training data.
However, their performance depends strongly on predefined assumptions, which may become unreliable in complex backgrounds.
To reduce this dependence on handcrafted priors, data-driven methods learn target representations directly from annotated infrared images.

Data-driven ISTD methods mainly include single-frame and multi-frame approaches.
Single-frame methods, such as ACM, ALCNet, DNA-Net, ISNet, UIU-Net, MSHNet, SCTransNet, and L2SKNet, learn spatial target--background representations from individual infrared images~\cite{ACM,ALCNet,li2023dnanet,zhang2022isnet,wu2022uiu,liu2024mshnet,yuan2024sctransnet,wu2025l2sknet}.
These methods are relatively efficient and have achieved substantial improvements through deep spatial feature learning.
Nevertheless, single-frame methods independently process each frame and do not exploit temporal features across consecutive frames.
To incorporate such temporal features, multi-frame methods, including STDMANet, DTUM, SSTNet, Tridos, and DeepPro, process infrared image sequences~\cite{STDMANet2023,DTUM,chen2024sstnet,Duan_2024_Tridos,li2025deeppro}.
Although temporal information can improve target observability and detection robustness, multi-frame processing introduces additional computational costs and may depend on reliable temporal correspondence.
In parallel, recent vision--language methods, such as motion-aware vision--language models, SAIST, and SeViL, have introduced textual semantics or target priors into infrared target detection~\cite{chen2025motion,SAIST,duan2026sevil}.
However, these methods often require costly task-specific semantic annotations.
Moreover, the large gap between high-level language semantics and the weak, low-resolution responses of infrared small targets makes fine-grained vision--language alignment difficult.

Although the above methods exploit different spatial, temporal, or semantic information, most learning-based detectors are ultimately optimized using classification and localization losses derived from bounding-box annotations.
These annotations define the final target location and extent, but they do not describe how ambiguous candidate regions should be ranked before localization, as illustrated in Fig.~\ref{fig:motivation}(b).
As a result, candidate-region prioritization can only be learned indirectly through the final detection objective.
This indirect learning may become unreliable when clutter or target-like background structures produce stronger responses than the true weak target.
This limitation motivates the central question of this work:
\emph{Can a dense spatial priority signal provide pre-localization guidance that complements bounding-box localization supervision for ISTD?}

% Although the above methods exploit different spatial, temporal, or semantic information, most learning-based detectors are still optimized mainly by classification and localization losses derived from bounding-box annotations.
% Bounding-box annotations specify the final target locations and extents, but they do not explicitly indicate the relative priority of ambiguous candidate regions before localization, as illustrated in Fig.~\ref{fig:motivation}(b).
% % Consequently, the detector must learn candidate-region priority implicitly from the final detection objective.
% Consequently, the final detection objective provides only implicit supervision for candidate-region prioritization.
% Such implicit guidance may be insufficient when background structures produce stronger target-like responses than the true weak target.
% Therefore, an important question arises:
% \emph{Can a dense spatial priority signal provide pre-localization guidance that complements bounding-box localization supervision for ISTD?}

To obtain such pre-localization guidance, task-driven human visual search offers a principled reference.
When observers search for a specific target, they do not examine all image locations equally.
Instead, top-down guidance related to the search goal and visual evidence from the current scene jointly influence where attention is directed~\cite{najemnik2005optimal,torralba2006contextual,fecteau2006priority}.
According to Guided Search 6.0~\cite{wolfe2021guided_search}, these guidance sources jointly shape a dynamic spatial \emph{priority map} that ranks candidate locations and makes high-priority locations more likely to be examined first~\cite{wolfe2021guided_search}.
Unlike classical bottom-up saliency, which is mainly determined by visual feature contrast, spatial priority also reflects target relevance and task-related information~\cite{itti1998saliency,fecteau2006priority,wolfe2021guided_search}.
This distinction is particularly relevant to ISTD because a strong clutter region may be visually salient but irrelevant to the target-search task, whereas a weak UAV may have low saliency but high task relevance.

Building on this priority-based selection principle, we propose Gaze-DETR, a bio-inspired priority-guided detector that predicts an internal priority map from infrared image features.
First, a priority head predicts a normalized priority map, in which higher values indicate regions that are more relevant to the weak-target search task.
Second, Residual Priority-Guided Feature Modulation (RPFM) uses the predicted map to modulate responses at high-priority locations while retaining the original multi-scale features, thereby helping preserve weak-target evidence.
Finally, Priority-Guided Anchor Query Injection (PAQI) converts high-priority locations into additional decoder anchor queries, guiding candidate selection and localization toward task-relevant regions.
Together, these components connect the learned spatial priority representation with both multi-scale feature processing and decoder candidate localization.

To provide priority-supervision maps for the predicted priority map under different annotation conditions, we introduce three priority-supervision schemes.
As illustrated in Fig.~\ref{fig:motivation}(c), each scheme adds a dense priority loss during training while retaining conventional bounding-box supervision.
The first scheme generates box-derived priority-supervision maps as Gaussian distributions directly from training bounding boxes and therefore requires no additional eye-tracking annotations.
The second scheme constructs real-gaze priority-supervision maps from fixation-density maps derived from task-driven eye tracking.
The third scheme models the statistical gaze--box relation from paired real-gaze and bounding-box annotations and then applies the learned relation to the Anti-UAV410 training boxes to generate transferred pseudo-gaze priority-supervision maps.
Thus, the three schemes differ only in how the priority-supervision map is constructed, while the Gaze-DETR architecture and the definition of the predicted priority map remain unchanged.

To support both real-gaze supervision and the gaze--box relation modeling required for pseudo-gaze transfer, we construct TIR-UAV120-Gaze with paired thermal infrared detection and task-driven eye-tracking annotations.
On this benchmark, fixation-density maps record the spatial distribution of gaze during weak-small UAV search, while bounding boxes provide final target-localization supervision.
Across the two benchmarks, all box-derived, real-gaze, and transferred pseudo-gaze priority-supervision maps are used only during training.
Therefore, Gaze-DETR produces the predicted priority map from the infrared image alone and maintains image-only inference on both datasets.

Extensive experiments evaluate Gaze-DETR under the three priority-supervision schemes.
As shown in Fig.~\ref{fig:performance_gain_over_dino}, all evaluated priority-supervision settings improve the DINO-DETR baseline on their corresponding benchmarks.
On TIR-UAV120-Gaze, Gaze-DETR achieves 85.76 mAP$_{50}$ and 88.77 F1 with box-derived priority supervision, and 86.18 mAP$_{50}$ and 89.00 F1 with real-gaze priority supervision.
On Anti-UAV410, where real eye-tracking annotations are unavailable, Gaze-DETR achieves 87.06 mAP$_{50}$ and 90.90 F1 with box-derived priority supervision, and 87.08 mAP$_{50}$ and 90.43 F1 with transferred pseudo-gaze priority supervision.
Taken together, these results demonstrate that explicit spatial-priority learning provides effective pre-localization guidance that complements conventional bounding-box localization supervision under different annotation conditions and costs.

The main contributions of this work are summarized as follows.
\begin{itemize}
    \item We formulate pre-localization candidate-region prioritization as a complementary supervision problem for box-supervised ISTD.
    Based on task-driven visual search, we introduce explicit spatial-priority learning to guide candidate processing before final target localization.

    \item We propose Gaze-DETR, a bio-inspired priority-guided detector that predicts an internal priority map from infrared images.
    The predicted map is used by RPFM for priority-guided multi-scale feature modulation and by PAQI for priority-guided anchor-query injection, while the complete detector remains image-only at inference.

    \item We introduce three priority-supervision schemes for different annotation conditions: box-derived Gaussian supervision, real-gaze fixation-density supervision, and transferred pseudo-gaze supervision based on statistical gaze--box relation modeling.
    To support real-gaze supervision and pseudo-gaze transfer, we construct TIR-UAV120-Gaze with paired thermal infrared detection and task-driven eye-tracking annotations.
    Experiments on TIR-UAV120-Gaze and Anti-UAV410 validate the proposed formulation under different annotation conditions and costs.
\end{itemize}
\section{Related Work}
\label{sec:related_work}

\subsection{Infrared Small-Target Detection}

Existing ISTD methods can be broadly divided into model-driven and data-driven approaches.
Model-driven methods distinguish targets from backgrounds using handcrafted operators or predefined structural priors.
As a classical approach, the top-hat transformation enhances compact bright structures by subtracting a morphologically estimated background~\cite{Tophat}.
Low-rank and sparse models instead represent targets as sparse components against structured backgrounds and extend this assumption to patch tensors or spatial--temporal tensors~\cite{gao2013infrared,dai2017reweighted,Liu2020STT}.
Other representative methods include max-mean filtering and local-contrast modeling~\cite{MaxMeandeshpande1999max,chen2013lcm}.
Although these methods are interpretable and require no large-scale training set, their performance depends on predefined target--background assumptions that may become unreliable in complex scenes.
Therefore, data-driven approaches have been developed to learn target representations directly from annotated infrared data.

According to the input information used, data-driven ISTD methods can be further divided into single-frame and multi-frame approaches.
Single-frame methods learn spatial target--background representations from individual infrared images.
As an early representative method, ACM introduces asymmetric contextual modulation to combine target responses with surrounding context~\cite{ACM}.
More recently, L2SKNet improves spatial representation by adaptively modeling local structures at different scales~\cite{wu2025l2sknet}.
Other representative methods include ALCNet, DNA-Net, ISNet, UIU-Net, MSHNet, and SCTransNet~\cite{ALCNet,li2023dnanet,zhang2022isnet,wu2022uiu,liu2024mshnet,yuan2024sctransnet}.
These methods learn more flexible spatial representations than model-driven approaches.
However, because they process each infrared image independently, they do not exploit temporal features across consecutive frames.

To incorporate temporal features, multi-frame methods process infrared image sequences.
STDMANet models spatial--temporal differences at multiple scales, whereas DeepPro explores long-term temporal-profile information~\cite{STDMANet2023,li2025deeppro}.
Other representative methods include DTUM, SSTNet, and Tridos~\cite{DTUM,chen2024sstnet,Duan_2024_Tridos}.
By exploiting cross-frame differences, motion patterns, or temporal correlations, these methods complement spatial features with temporal information.
Nevertheless, multi-frame processing introduces additional computational and memory costs and may depend on reliable temporal correspondence.

In parallel, foundation-model and vision--language methods introduce pretrained knowledge or textual semantics into ISTD.
For example, SAIST transfers contrastive vision--language knowledge to infrared small-target segmentation, while SeViL introduces semantic prompts into semi-supervised moving infrared small-target detection~\cite{SAIST,duan2026sevil}.
Other related approaches include IRSAM and motion-aware vision--language learning~\cite{zhang2024irsam,chen2025motion}.
These methods supplement weak infrared appearance with pretrained or semantic knowledge.
However, task-specific semantic annotations can be costly to obtain, and the gap between high-level language semantics and weak, low-resolution target responses makes fine-grained cross-modal alignment difficult.

Overall, existing learning-based methods mainly improve spatial, temporal, or semantic representations.
However, their classification and localization losses primarily supervise final detection results and do not explicitly specify the relative priority of ambiguous candidate regions before localization.
Consequently, candidate-region priority is generally learned only implicitly from the final detection objective.
This limitation motivates the use of an additional spatial priority signal to complement conventional bounding-box supervision.

\subsection{Task-Driven Visual Search and Gaze Supervision}

Research on visual attention initially focused on bottom-up saliency.
The classical model of Itti et al.~\cite{itti1998saliency} identifies visually distinctive regions according to local feature contrast.
Subsequent image and video saliency methods learn general fixation distributions under free-viewing conditions~\cite{judd2012mit300,huang2015salicon,wang2018dhf1k,kummerer2022deepgaze}.
However, free-viewing saliency mainly describes image-driven visual conspicuity and does not fully represent attention allocation during a specific target-search task.

In contrast, task-driven visual search is jointly influenced by visual evidence and knowledge about the search target.
Optimal-search and contextual-guidance studies show that target information and scene context affect fixation selection~\cite{najemnik2005optimal,torralba2006contextual}.
Guided Search 6.0 further explains that these guidance sources contribute to a spatial priority map that ranks candidate locations according to their relevance to the current task~\cite{fecteau2006priority,wolfe2021guided_search}.
Therefore, spatial priority reflects not only visual contrast but also target relevance.
This distinction is important for ISTD, where strong clutter may be visually salient but irrelevant to UAV search, whereas a weak UAV may remain highly task-relevant.

Eye tracking provides spatial information about attention allocation during visual search.
Fixation-density maps aggregate gaze locations and describe the regions in which observers concentrate their attention~\cite{yarbus1967eye,holmqvist2011eye_tracking}.
As a representative task-driven dataset, COCO-Search18 records fixation sequences while observers search for specified objects in natural scenes~\cite{chen2021cocosearch18}.
Other eye-tracking studies have considered medical-image analysis, attention prediction, remote-sensing images, and UAV videos~\cite{saab2021observational,zhou2024observergaze,li2024eeg_eyetracking_rs,perrin2020eyetrackuav2}.
Together, these studies show that gaze can provide dense spatial information beyond conventional category or localization annotations.
Nevertheless, the use of task-driven fixation-density maps to supervise a predicted priority map for infrared weak-small UAV detection remains insufficiently explored.

\subsection{DETR-Based Detection and Priority Guidance}

DETR-based detectors formulate object detection as an end-to-end set-prediction problem.
DETR uses learnable object queries and a Transformer decoder to predict object categories and bounding boxes~\cite{carion2020end}.
Deformable DETR subsequently improves multi-scale feature processing through sparse deformable attention, and DINO further strengthens query initialization and denoising training~\cite{zhu2021deformable,zhang2023dino}.
Other query-based detectors, such as RT-DETR, improve inference efficiency~\cite{zhao2024rtdetr}.
Together, these developments provide a unified framework for candidate selection and target localization.

However, existing DETR-based detectors mainly learn object queries and encoder proposals from final classification and localization losses.
Therefore, the spatial priority of ambiguous candidate regions is still learned implicitly.
For infrared weak-small targets, strong clutter responses may consequently dominate candidate selection before the true weak target is localized.
Different from these methods, Gaze-DETR learns a predicted priority map through explicit priority supervision and uses it for both multi-scale feature processing and anchor-query injection.
Because all priority-supervision maps are used only during training, the detector retains image-only inference.

\section{Method}
\label{sec:method}

\subsection{Problem Formulation and Framework Overview}
\label{sec:method_overview}

Gaze-DETR is built on the DINO-DETR framework~\cite{zhang2023dino}.
Given an infrared image $\mathbf{I}_i$, the detector predicts a set of object candidates:
\begin{equation}
\hat{\mathcal{Y}}_i
=
\left\{
\left(
\hat{\mathbf{b}}_{ij},
\hat{\mathbf{p}}_{ij}
\right)
\right\}_{j=1}^{N_q},
\end{equation}
where $\hat{\mathbf{b}}_{ij}$ denotes the bounding box predicted by the $j$-th query, $\hat{\mathbf{p}}_{ij}$ denotes its class-probability vector, and $N_q$ is the total number of decoder queries.
The corresponding ground-truth set is written as
\begin{equation}
\mathcal{Y}_i
=
\left\{
\left(
\mathbf{b}_{ik},
c_{ik}
\right)
\right\}_{k=1}^{N_i},
\end{equation}
where $\mathbf{b}_{ik}$ and $c_{ik}$ denote the bounding box and category of the $k$-th target, respectively, and $N_i=0$ for a target-absent frame.
Following DINO-DETR, classification and localization losses supervise the final detection results.

Although the standard detection pathway predicts the final target location, it does not explicitly represent the relative spatial priority of candidate regions before localization.
To provide this intermediate guidance, Gaze-DETR introduces a priority pathway that predicts a dense priority map from image features.
Let
\begin{equation}
\hat{\mathbf{P}}_i
\in
[0,1]^{H_p\times W_p}
\end{equation}
denote the predicted priority map, where $H_p=W_p=64$ in our implementation.
The map is spatially normalized:
\begin{equation}
\sum_{u=1}^{H_p}
\sum_{v=1}^{W_p}
\hat{\mathbf{P}}_i(u,v)
=1.
\end{equation}
Accordingly, a larger value indicates that the corresponding location receives a higher relative priority for weak-target search.
The map represents a spatial ranking of candidate locations rather than the probability that a target is present in the image.

During training, a dense priority-supervision map $\mathbf{G}_i$ and a sample-specific weight $w_i^{pri}$ supervise the predicted priority map $\hat{\mathbf{P}}_i$.
Depending on the scheme, $\mathbf{G}_i$ is a box-derived, real-gaze, or transferred pseudo-gaze priority-supervision map, while $w_i^{pri}$ controls its contribution to the priority loss.
The three schemes for constructing the priority-supervision map are detailed in Section~\ref{sec:gaze_target_construction}.
In the remainder of this paper, \emph{predicted priority map} refers to the detector output $\hat{\mathbf{P}}_i$, whereas \emph{priority-supervision map} refers to the training-only spatial map $\mathbf{G}_i$ used to supervise that output. A \emph{fixation-density map} is the observation-derived gaze representation used to construct the real-gaze priority-supervision map.
Importantly, $\mathbf{G}_i$ is used only to calculate the priority loss.
RPFM and PAQI never directly access this priority-supervision map; instead, both modules operate on the predicted priority map $\hat{\mathbf{P}}_i$.

\begin{figure*}[t]
    \centering
    \makebox[\textwidth][c]{%
        \hspace*{4mm}%
        \includegraphics[
            width=0.85\textwidth,
            trim={0pt 0pt 0pt 0pt},
            clip
        ]{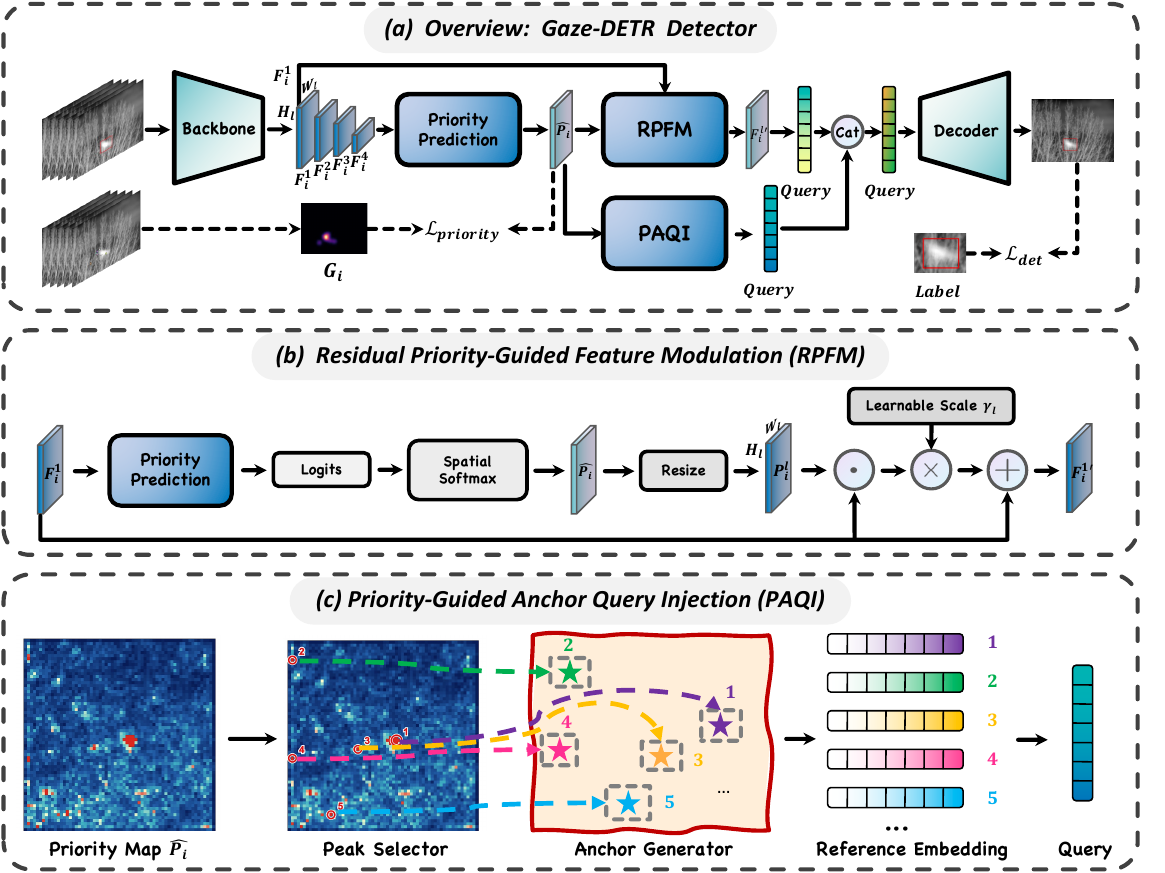}%
    }
    \caption{Overview and main components of Gaze-DETR.
    (a) The backbone extracts multi-scale image features, from which the priority head produces the predicted priority map $\hat{\mathbf{P}}_i$.
    During training, $\mathcal{L}_{priority}$ aligns the predicted priority map with the priority-supervision map $\mathbf{G}_i$.
    RPFM uses the predicted map for residual modulation of multi-scale features, whereas PAQI converts high-priority locations into additional decoder anchor queries.
    (b) RPFM introduces priority-dependent residual modulation while preserving the original feature pathway.
    (c) PAQI selects spatially separated priority peaks and combines them with weak-target size priors to construct additional anchor queries.
    During inference, the detector requires only the infrared image and does not use gaze annotations, priority-supervision maps, or test bounding boxes.}
    \label{fig:method_overview}
\end{figure*}

Figure~\ref{fig:method_overview}(a) summarizes the complete framework.
First, the backbone and projection layers extract a multi-scale feature pyramid from the infrared image.
The priority head then predicts $\hat{\mathbf{P}}_i$ from the highest-resolution projected feature.
Next, RPFM introduces the predicted spatial priority into multi-scale feature processing.
Finally, PAQI converts high-priority locations into additional anchor queries for decoder localization.
Thus, the same internal priority representation connects feature processing with candidate localization.

The complete architecture, including the priority head, RPFM, and PAQI, is used in all reported experiments.
Across datasets and supervision settings, only the construction of $\mathbf{G}_i$ changes; the network architecture and image-only inference protocol remain unchanged.
We next describe priority prediction, feature-level modulation, query-level injection, and the unified training objective in this order.

\subsection{Visual Encoding and Priority Prediction}
\label{sec:priority_prediction}

We first describe how Gaze-DETR produces the predicted priority map from the infrared image.
The DINO-DETR backbone and projection layers produce a multi-scale feature pyramid:
\begin{equation}
\mathcal{F}_i
=
\left\{
\mathbf{F}_i^l
\right\}_{l=1}^{L},
\qquad
\mathbf{F}_i^l
\in
\mathbb{R}^{C\times H_l\times W_l},
\end{equation}
where $L$ is the number of feature levels and $C$ is the projected channel dimension.
The original DINO encoder, decoder, denoising training, and Hungarian matching strategy are retained.

The priority head $\Psi(\cdot)$ takes the highest-resolution projected feature as input:
\begin{equation}
\mathbf{Z}_i
=
\Psi
\left(
\mathbf{F}_i^1
\right),
\end{equation}
where
$\mathbf{Z}_i\in\mathbb{R}^{H_p\times W_p}$
denotes the spatial priority logits.
In our implementation, $\Psi$ contains a $3\times3$ convolution with 256 input and output channels, GroupNorm with 32 groups, ReLU activation, and a $1\times1$ convolution that produces one logit channel.
The resulting logits are bilinearly resized to $64\times64$.

A spatial softmax converts the logits into the normalized priority map:
\begin{equation}
\hat{\mathbf{P}}_i(u,v)
=
\frac{
\exp\left(\mathbf{Z}_i(u,v)\right)
}{
\sum_{u',v'}
\exp\left(\mathbf{Z}_i(u',v')\right)
}.
\end{equation}
Because this map is predicted entirely from image features, the same priority-prediction pathway is used during both training and inference.

For each valid priority-supervision map, we supervise the predicted priority map using a combination of Kullback--Leibler divergence and histogram-intersection similarity:
\begin{equation}
\mathcal{L}_{priority}^{i}
=
\operatorname{KL}
\left(
\mathbf{G}_i
\|
\hat{\mathbf{P}}_i
\right)
+
\lambda_{sim}
\left[
1-
\operatorname{SIM}
\left(
\mathbf{G}_i,
\hat{\mathbf{P}}_i
\right)
\right].
\end{equation}

The KL-divergence term is defined as
\begin{equation}
\operatorname{KL}
\left(
\mathbf{G}_i
\|
\hat{\mathbf{P}}_i
\right)
=
\sum_{u,v}
\mathbf{G}_i(u,v)
\log
\frac{
\mathbf{G}_i(u,v)+\epsilon
}{
\hat{\mathbf{P}}_i(u,v)+\epsilon
}.
\end{equation}
This term penalizes differences between the target and predicted spatial distributions.

The similarity term is defined as
\begin{equation}
\operatorname{SIM}
\left(
\mathbf{G}_i,
\hat{\mathbf{P}}_i
\right)
=
\sum_{u,v}
\min
\left[
\mathbf{G}_i(u,v),
\hat{\mathbf{P}}_i(u,v)
\right].
\end{equation}
Whereas the KL term measures distributional discrepancy, the SIM term directly rewards the probability mass shared by the two maps.
Therefore, the two terms provide complementary distribution-level supervision.

The weighted priority loss over a mini-batch is
\begin{equation}
\mathcal{L}_{priority}
=
\frac{
\sum_i
w_i^{pri}
\mathcal{L}_{priority}^{i}
}{
\sum_i
w_i^{pri}
+\epsilon
}.
\end{equation}
If a mini-batch contains no valid priority-supervision map, we set
$\mathcal{L}_{priority}=0$.
Once predicted, the predicted priority map is used by RPFM at the feature level and by PAQI at the query level.

\subsection{Residual Priority-Guided Feature Modulation}
\label{sec:feature_modulation}

RPFM introduces the predicted priority map into multi-scale feature processing, as illustrated in Fig.~\ref{fig:method_overview}(b).
For each feature level $l$, the predicted priority map is first resized to the corresponding spatial resolution:
\begin{equation}
\hat{\mathbf{P}}_i^l
=
\operatorname{Resize}
\left(
\hat{\mathbf{P}}_i,
H_l,
W_l
\right).
\end{equation}

The resized map is then detached from the detection-gradient pathway, broadcast along the channel dimension, and used for residual feature modulation:
\begin{equation}
\tilde{\mathbf{F}}_i^l
=
\mathbf{F}_i^l
+
\gamma_l
\left(
\operatorname{sg}\!\left(\hat{\mathbf{P}}_i^l\right)
\odot
\mathbf{F}_i^l
\right),
\end{equation}
where $\odot$ denotes element-wise multiplication,
$\operatorname{sg}(\cdot)$ denotes the stop-gradient operation, and
$\gamma_l$ is a learnable scalar for the $l$-th feature level.

Each $\gamma_l$ is initialized to zero.
Consequently, the initial RPFM output is identical to the original DINO feature, allowing optimization to start from the pretrained detector behavior.
During training, $\gamma_l$ gradually controls the magnitude and direction of priority-dependent modulation at each feature level.

The residual formulation preserves the complete original feature pathway.
Therefore, RPFM neither hard-masks low-priority locations nor removes their visual information.
Instead, it introduces priority-dependent changes while retaining the original multi-scale representation.

The resulting feature pyramid is
\begin{equation}
\tilde{\mathcal{F}}_i
=
\left\{
\tilde{\mathbf{F}}_i^l
\right\}_{l=1}^{L}.
\end{equation}
This modulated pyramid is subsequently processed by the transformer encoder.
While RPFM uses spatial priority to modify feature processing, PAQI uses the same predicted representation to guide decoder-query initialization.

\subsection{Priority-Guided Anchor Query Injection}
\label{sec:priority_anchor_query}

PAQI converts high-priority locations into additional spatial anchors for decoder localization, as illustrated in Fig.~\ref{fig:method_overview}(c).
Before peak selection, the predicted priority map is detached:
\begin{equation}
\mathbf{P}_i^{sg}
=
\operatorname{sg}
\left(
\hat{\mathbf{P}}_i
\right).
\end{equation}
Thus, the priority head is optimized directly by $\mathcal{L}_{priority}$, while the detection losses optimize the injected query embeddings and the downstream detector.

The original two-stage DINO pathway first produces detection-query contents and reference boxes:
\begin{equation}
\mathbf{Q}_i^{det}
=
\left\{
\mathbf{q}_{ij}^{det}
\right\}_{j=1}^{N_d},
\qquad
\mathbf{R}_i^{det}
=
\left\{
\mathbf{r}_{ij}^{det}
\right\}_{j=1}^{N_d},
\end{equation}
where $N_d$ denotes the number of original DINO queries.
All original detection queries are retained.

To construct additional candidate locations, a peak selector extracts $K$ spatially separated local maxima from $\mathbf{P}_i^{sg}$:
\begin{equation}
\mathcal{C}_i
=
\left\{
(x_{im},y_{im},s_{im})
\right\}_{m=1}^{K},
\end{equation}
where $(x_{im},y_{im})$ denotes the normalized position of the $m$-th peak and $s_{im}$ denotes its priority value.
The peaks are selected in descending order of priority.
After each selection, local suppression is applied to prevent multiple peaks from being concentrated in the same small neighborhood.

A selected peak determines a candidate location but does not define the target extent.
Therefore, each location is paired with a set of weak-target size priors:
\begin{equation}
\mathcal{S}
=
\left\{
(w_r,h_r)
\right\}_{r=1}^{R},
\end{equation}
where the normalized width--height pairs are estimated from the training-set bounding-box distribution.
Combining the selected centers with these size priors produces the priority-guided reference boxes:
\begin{equation}
\mathbf{r}_{imr}^{pri}
=
(x_{im},y_{im},w_r,h_r),
\end{equation}
where
\begin{equation}
m=1,\ldots,K,
\qquad
r=1,\ldots,R.
\end{equation}
The resulting number of priority-guided anchors is
\begin{equation}
N_p=KR.
\end{equation}

The corresponding query contents are represented by learnable embeddings:
\begin{equation}
\mathbf{Q}_i^{pri}
=
\left\{
\mathbf{e}_{a}^{pri}
+
\mathbf{e}_{type}^{pri}
\right\}_{a=1}^{N_p}.
\end{equation}
Thus, the predicted priority map determines where the additional queries are initialized, whereas the learnable embeddings provide their query contents.

Before decoder processing, the original and priority-guided queries are concatenated:
\begin{equation}
\mathbf{Q}_i^{all}
=
\left[
\mathbf{Q}_i^{det};
\mathbf{Q}_i^{pri}
\right],
\qquad
\mathbf{R}_i^{all}
=
\left[
\mathbf{R}_i^{det};
\mathbf{R}_i^{pri}
\right].
\end{equation}
The total number of decoder queries is therefore
\begin{equation}
N_q=N_d+N_p.
\end{equation}

All queries are processed by the same decoder and optimized through the same Hungarian matching and detection losses.
Consequently, PAQI does not require a separate prediction head or an additional detection rule.
Instead, it supplements the original DINO queries with candidates initialized at high-priority locations.

PAQI is activated after an initial warm-up period.
This schedule allows the priority head to learn a sufficiently stable spatial distribution before its local maxima are used for query initialization.
After the warm-up period, the same predicted-map-to-query procedure is applied during both training and inference.
Neither ground-truth priority-supervision maps nor bounding boxes are used to construct the additional queries at test time.

\subsection{Training Objective and Image-Only Inference}
\label{sec:training_objective}

The complete training objective combines the original DINO detection loss with the auxiliary priority loss:
\begin{equation}
\mathcal{L}
=
\mathcal{L}_{det}
+
\lambda_p
\mathcal{L}_{priority},
\end{equation}
where $\lambda_p$ controls the contribution of priority supervision.

The detection term follows DINO-DETR:
\begin{equation}
\mathcal{L}_{det}
=
\mathcal{L}_{cls}
+
\lambda_{box}\mathcal{L}_{1}
+
\lambda_{giou}\mathcal{L}_{giou}
+
\mathcal{L}_{dn}
+
\mathcal{L}_{aux},
\end{equation}
where $\mathcal{L}_{cls}$ denotes the classification loss,
$\mathcal{L}_{1}$ denotes the bounding-box regression loss,
$\mathcal{L}_{giou}$ denotes the generalized IoU loss~\cite{rezatofighi2019generalized},
$\mathcal{L}_{dn}$ denotes the denoising loss, and
$\mathcal{L}_{aux}$ contains the auxiliary encoder and decoder losses.

During training, the target $\mathbf{G}_i$ affects the network only through $\mathcal{L}_{priority}$.
It is never directly supplied to RPFM or PAQI.
Instead, both modules consume the predicted priority map $\hat{\mathbf{P}}_i$.
Therefore, the same internal priority representation is used by RPFM and PAQI during both training and inference.

At inference, the only external input is the infrared image $\mathbf{I}_i$.
The backbone and priority head first predict $\hat{\mathbf{P}}_i$ from the image.
RPFM then uses this map for residual multi-scale feature modulation, while PAQI uses it to initialize additional decoder anchor queries.
Raw gaze samples, gaze coordinates, fixation-density maps, transferred pseudo-gaze descriptors, priority-supervision maps, and test-set bounding boxes are not provided.
Therefore, Gaze-DETR retains the deployment interface of a standard image-only detector.
\section{Priority-Supervision Schemes}
\label{sec:priority_supervision_schemes}

This section first defines the three priority-supervision schemes used to train the shared priority head under different annotation conditions.
It then introduces TIR-UAV120-Gaze, which provides the paired detection and task-driven eye-tracking annotations required for real-gaze supervision and gaze--box relation modeling.

\subsection{Schemes Themes}
\label{sec:gaze_target_construction}

Under the unified architecture in Section~\ref{sec:method}, the three schemes differ only in how the priority-supervision map is constructed.
The box-derived scheme generates a Gaussian map directly from an existing training bounding box and therefore requires no additional eye-tracking annotation.
When paired eye-tracking annotations are available, the real-gaze scheme constructs the map from a task-driven fixation-density map.
When real gaze is unavailable, the transferred pseudo-gaze scheme first models the statistical gaze--box relation from paired annotations in TIR-UAV120-Gaze and then applies this relation to the Anti-UAV410 training boxes.
Accordingly, the priority-supervision map is instantiated as $\mathbf{G}_i^{box}$, $\mathbf{G}_i^{real}$, or $\mathbf{G}_i^{pseudo}$, while the Gaze-DETR architecture remains unchanged.
The benchmark that supports real-gaze supervision and gaze--box relation modeling is introduced next in Section~\ref{sec:benchmark_construction}.

A sample-specific weight $w_i^{pri}$ determines whether and to what extent a priority-supervision map contributes to the priority loss.
For a valid box-derived or real-gaze priority-supervision map, $w_i^{pri}=1$.
For an invalid gaze annotation or a target-absent frame, $w_i^{pri}=0$.
For a transferred pseudo-gaze priority-supervision map, the weight is determined by the reliability transferred from the real-gaze training data.

\subsubsection{Box-Derived Priority-Supervision Map}

The box-derived priority-supervision map converts a sparse localization annotation into a dense spatial distribution.
For a target-present frame, let $(c_i^x,c_i^y)$ denote the center of the training bounding box.
After mapping this center to the $H_p\times W_p$ priority grid, we obtain
$(\bar{c}_i^x,\bar{c}_i^y)$.
Here, $u$ and $v$ denote the row and column coordinates of the priority-supervision map, respectively.

The box-derived priority-supervision map is defined as
\begin{equation}
\mathbf{G}_i^{box}(u,v)
=
\frac{1}{Z_i^{box}}
\exp
\left[
-\frac{(v-\bar{c}_i^x)^2}{2(\sigma_{b,i}^{x})^2}
-\frac{(u-\bar{c}_i^y)^2}{2(\sigma_{b,i}^{y})^2}
\right],
\end{equation}
where the horizontal and vertical spreads are determined by the size of the training box:
\begin{equation}
\begin{aligned}
\sigma_{b,i}^{x}
&=
\max\left(1.5,0.5W_p\frac{w_i^b}{W_i}\right),\\
\sigma_{b,i}^{y}
&=
\max\left(1.5,0.5H_p\frac{h_i^b}{H_i}\right).
\end{aligned}
\end{equation}
Here, $(w_i^b,h_i^b)$ denotes the bounding-box size, while $(W_i,H_i)$ denotes the image size.
Therefore, $\mathbf{G}_i^{box}$ is a size-adaptive anisotropic Gaussian rather than a fixed isotropic distribution.

The normalization factor $Z_i^{box}$ ensures that
\begin{equation}
\sum_{u,v}
\mathbf{G}_i^{box}(u,v)
=1.
\end{equation}
Because this priority-supervision map is derived from an existing training box, it introduces no additional annotation cost.
At the same time, its dense spatial form allows us to evaluate whether explicit priority supervision provides complementary information beyond the standard classification and localization losses.

\subsubsection{Real-Gaze Priority-Supervision Map}

When real eye-tracking annotations are available, the real-gaze priority-supervision map is constructed from the frame-aligned fixation-density map.
To maintain spatial correspondence, the fixation-density map undergoes the same geometric transformation as the infrared image.
The transformed map is then resized to the priority-grid resolution:
\begin{equation}
\tilde{\mathbf{M}}_i
=
\operatorname{Resize}
\left(
\mathcal{T}_i(\mathbf{M}_i),
H_p,
W_p
\right),
\end{equation}
where $\mathbf{M}_i$ denotes the original fixation-density map and $\mathcal{T}_i$ denotes the geometric transformation applied during training.

After possible negative numerical values are removed, the real-gaze priority-supervision map is normalized as
\begin{equation}
\mathbf{G}_i^{real}(u,v)
=
\frac{
\max\left(\tilde{\mathbf{M}}_i(u,v),0\right)
}{
\sum_{u,v}
\max\left(\tilde{\mathbf{M}}_i(u,v),0\right)
+\epsilon
}.
\end{equation}
This normalization preserves the relative spatial concentration of gaze while making the priority-supervision map compatible with the predicted priority map.

In a small number of cases, a valid gaze coordinate is available but the corresponding density map cannot be used.
For these samples, the transformed gaze coordinate is mapped to the priority grid and rendered as
\begin{equation}
\mathbf{G}_i^{real}(u,v)
=
\frac{1}{Z_i^{real}}
\exp
\left[
-\frac{
(v-\bar{x}_i)^2+
(u-\bar{y}_i)^2
}{
2\sigma_g^2
}
\right],
\end{equation}
where $(\bar{x}_i,\bar{y}_i)$ denotes the transformed gaze coordinate, $\sigma_g$ controls the spatial spread, and $Z_i^{real}$ normalizes the map.

The resulting real-gaze priority-supervision map retains the spatial concentration observed during task-driven UAV search.
Rather than reproducing fixation order or predicting a human scanpath, it provides dense supervision for learning the detector-side spatial priority representation.

Target-absent frames remain part of conventional detection training because they contribute to learning the no-object decision.
However, because these frames contain no visible target region for positive priority supervision, their priority weight is set to zero.

\subsubsection{Transferred Pseudo-Gaze Priority-Supervision Map}

Anti-UAV410 does not contain real eye-tracking annotations.
To obtain transferred pseudo-gaze priority supervision without collecting additional gaze data, we transfer the empirical relationship between gaze locations and target boxes from the TIR-UAV120-Gaze training split.
This procedure uses only training data and does not access Anti-UAV410 test boxes during model training or inference.

\begin{figure*}[t]
    \centering
    \includegraphics[
        width=0.98\textwidth,
        trim={0pt 0pt 0pt 0pt},
        clip
    ]{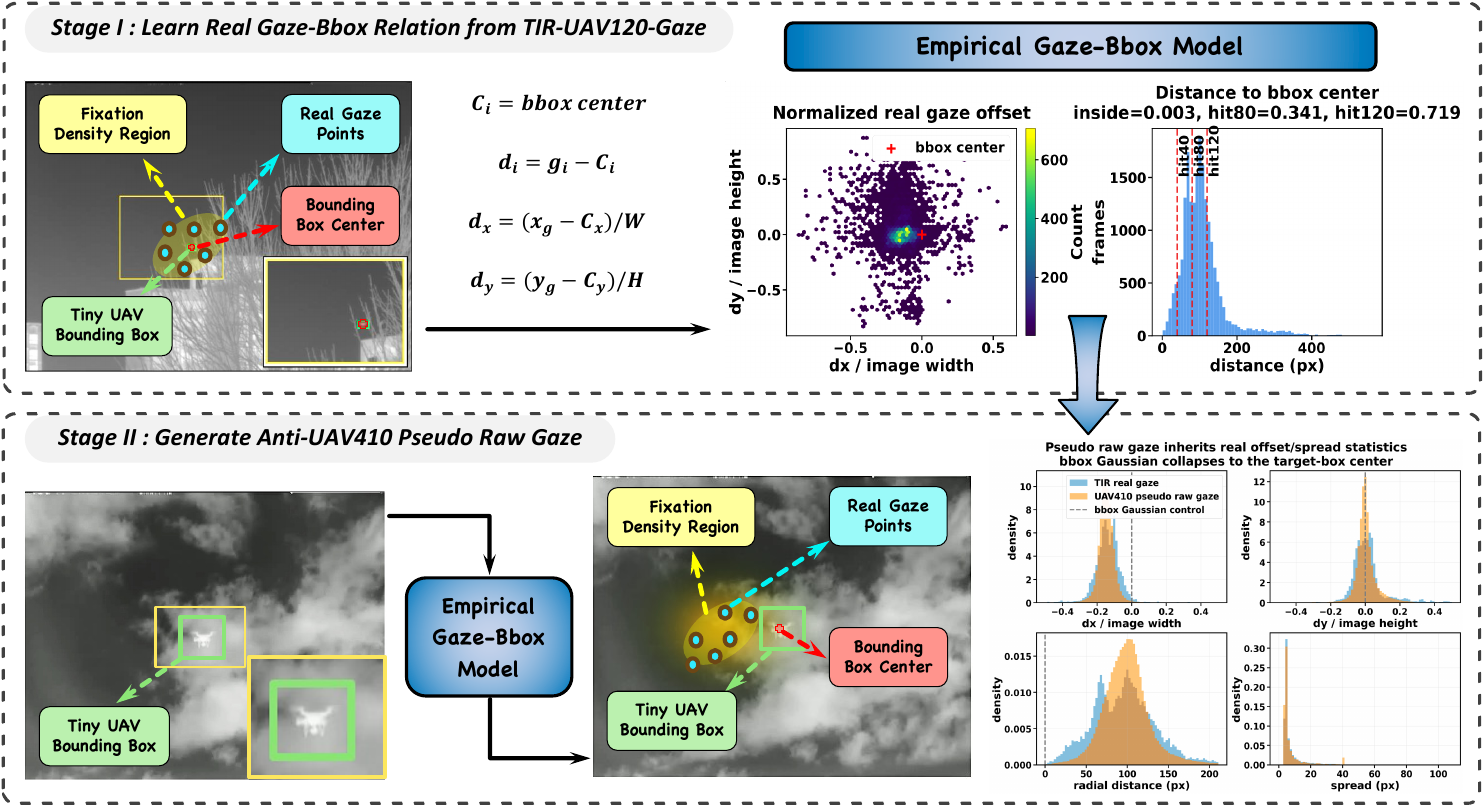}
    \caption{Construction of transferred pseudo-gaze priority-supervision maps for Anti-UAV410.
    Stage I estimates box-conditioned gaze-offset and reliability distributions from the TIR-UAV120-Gaze training split.
    Stage II samples a pseudo-gaze center for each Anti-UAV410 training box, applies within-video temporal smoothing, and renders the resulting center as a Gaussian priority-supervision map.
    The generated maps supervise only the priority head during training and are not used as detector inputs during inference.}
    \label{fig:antiuav410_pseudo_raw_gaze}
    \vspace{-2mm}
\end{figure*}

For a valid TIR-UAV120-Gaze training sample, let the image size be $(W_i,H_i)$ and the target box be
\begin{equation}
\mathbf{b}_i
=
(x_i^b,y_i^b,w_i^b,h_i^b).
\end{equation}
Its center is denoted by
$\mathbf{c}_i=(c_i^x,c_i^y)$.
The corresponding gaze descriptor is written as
\begin{equation}
\mathbf{g}_i
=
(x_i^g,y_i^g,v_i,s_i),
\end{equation}
where $(x_i^g,y_i^g)$ denotes the gaze coordinate, $v_i$ denotes the validity flag, and $s_i$ denotes the gaze strength.

To describe gaze relative to the target rather than to the absolute image coordinates, we define the normalized offset
\begin{equation}
\boldsymbol{\Delta}_i
=
\left(
\frac{x_i^g-c_i^x}{W_i},
\frac{y_i^g-c_i^y}{H_i}
\right).
\end{equation}
The corresponding reliability is defined as
\begin{equation}
r_i
=
\operatorname{clip}
\left(
v_i s_i,
0,
1
\right).
\end{equation}

We group valid training observations according to target-area intervals and construct empirical pools of
$\boldsymbol{\Delta}_i$ and $r_i$.
This grouping allows the transferred distribution to account for the relationship between target size and gaze location.
When a size-specific pool contains insufficient observations, the global training pool is used as a fallback.

For an Anti-UAV410 training frame at time $t$, an offset
$\boldsymbol{\Delta}_t$ and a reliability value $r_t$ are sampled from the corresponding empirical pool.
Because adjacent frames within the same video are temporally related, the sampled offsets are smoothed within each sequence:
\begin{equation}
\tilde{\boldsymbol{\Delta}}_t
=
\rho
\tilde{\boldsymbol{\Delta}}_{t-1}
+
(1-\rho)
\boldsymbol{\Delta}_t,
\end{equation}
where $\rho=0.45$.
For the first eligible frame of a sequence, the smoothed offset is initialized as
$\tilde{\boldsymbol{\Delta}}_t=\boldsymbol{\Delta}_t$.
The temporal state is propagated only when the gap from the preceding eligible frame in the same sequence is no greater than five frames; otherwise, the state is reinitialized.

The pseudo-gaze center is then calculated from the current training box:
\begin{equation}
\tilde{\mathbf{g}}_t^{xy}
=
\left(
c_t^x+W_t\tilde{\Delta}_t^x,
c_t^y+H_t\tilde{\Delta}_t^y
\right).
\end{equation}
Afterward, the coordinate is clipped to the image boundary and mapped to the priority grid.

The transferred pseudo-gaze priority-supervision map is rendered as
\begin{equation}
\mathbf{G}_t^{pseudo}(u,v)
=
\frac{1}{Z_t^{pseudo}}
\exp
\left[
-\frac{
(v-\bar{\tilde{x}}_t)^2+
(u-\bar{\tilde{y}}_t)^2
}{
2\sigma_p^2
}
\right],
\end{equation}
where
$(\bar{\tilde{x}}_t,\bar{\tilde{y}}_t)$ denotes the pseudo-gaze center on the priority grid,
$\sigma_p=8$, and
$Z_t^{pseudo}$ normalizes the map.

The sampled reliability does not change the amplitude of the normalized spatial distribution.
Instead, it controls the contribution of the sample to the priority loss:
\begin{equation}
w_t^{pri}
=
\operatorname{clip}
\left(
r_t,
0,
1
\right).
\end{equation}
For a target-absent frame, $w_t^{pri}=0$.
Therefore, the transferred pseudo-gaze priority-supervision map affects the detector only through priority-head supervision and is never supplied to RPFM, PAQI, or the inference pipeline.
Having defined the three supervision schemes, we next introduce the benchmark that provides the paired annotations required by the real-gaze and transfer schemes.

\subsection{TIR-UAV120-Gaze Benchmark}
\label{sec:benchmark_construction}

\subsubsection{Benchmark Motivation and Overview}
\label{sec:benchmark_overview}

Existing thermal infrared UAV benchmarks mainly provide localization annotations, such as bounding boxes or target trajectories, but they do not contain synchronized task-driven eye-tracking annotations.
In contrast, existing gaze and saliency datasets provide human attention data, but they are generally designed for natural-image free viewing, dynamic saliency prediction, goal-directed search in RGB scenes, or RGB UAV observation.
Therefore, neither type of resource directly supports the study of task-driven gaze supervision for thermal infrared weak-small UAV detection.

To address this limitation, we construct TIR-UAV120-Gaze, a benchmark that combines thermal infrared UAV videos, standard detection annotations, and task-driven eye-tracking annotations.
Its name reflects its three defining components: thermal infrared imaging (TIR), 120 UAV video sequences, and gaze annotations collected during target search.
The benchmark supports conventional detection evaluation, construction of real-gaze priority-supervision maps, and statistical gaze--box relation modeling for transferred pseudo-gaze priority-supervision maps.

TIR-UAV120-Gaze contains 120 thermal infrared UAV videos and 30,362 sampled frames at a resolution of $640{\times}480$.
Each sampled frame is associated with a detection annotation, a frame-level gaze descriptor, and a fixation-density map.
The detection annotation defines the target location for conventional detection supervision and evaluation.
Meanwhile, the gaze descriptor and fixation-density map record the spatial allocation of attention during the weak-small UAV search task.
Together, these annotations support final target localization, construction of real-gaze priority-supervision maps, and gaze--box relation modeling under a unified data protocol.

\begin{table*}[t]
\centering
\PaperTableFont
\setlength{\tabcolsep}{3.0pt}
\renewcommand{\arraystretch}{1.10}
\caption{Capability comparison with representative gaze, saliency, and Anti-UAV benchmarks.
The comparison focuses on four properties related to this work: thermal infrared imagery, real gaze, task-driven observation, and weak-small-target detection.}
\label{tab:benchmark_core_advantages}
\begin{adjustbox}{max width=\textwidth}
\begin{tabular}{lccccccc}
\toprule
\textbf{Dataset} &
\textbf{Publication} &
\textbf{\# Samples} &
\textbf{Primary Setting} &
\textbf{Infrared} &
\textbf{Real Gaze} &
\textbf{Task-Driven} &
\textbf{Weak-/Small-Target Focus} \\
\midrule

MIT300~\cite{judd2012mit300,mit_tuebingen_datasets}
& 2012
& 300 images
& Natural-image free viewing
& \abno & \abyes & \abno & \abno \\

CAT2000~\cite{borji2015cat2000,mit_tuebingen_datasets}
& 2015
& 4,000 images
& Category-aware free viewing
& \abno & \abyes & \abno & \abno \\

SALICON~\cite{huang2015salicon}
& 2015
& MS COCO images
& Surrogate saliency
& \abno & \abno & \abno & \abno \\

DHF1K~\cite{wang2018dhf1k}
& 2018
& 1,000 videos
& Dynamic video free viewing
& \abno & \abyes & \abno & \abno \\

COCO-Search18~\cite{chen2021cocosearch18,yang2024hat}
& 2021
& 6,202 images
& Goal-directed object search
& \abno & \abyes & \abyes & \abno \\

EyeTrackUAV2~\cite{perrin2020eyetrackuav2}
& 2020
& 43 RGB UAV videos
& UAV video observation
& \abno & \abyes & \abno & \abno \\

Anti-UAV300~\cite{jiang2021antiuav}
& 2021
& $>$300 video pairs
& RGB/IR Anti-UAV tracking
& \textit{partial} & \abno & \abno & \abyes \\

Anti-UAV410~\cite{huang2024antiuav410}
& 2024
& 410 TIR videos
& Thermal UAV tracking
& \abyes & \abno & \abno & \abyes \\

Anti-UAV600~\cite{zhu2023evidential}
& 2023
& 600 TIR videos
& Thermal UAV detection/tracking
& \abyes & \abno & \abno & \abyes \\

\rowcolor{orange!12}
\textbf{TIR-UAV120-Gaze (Ours)}
& \textbf{--}
& \textbf{120 videos / 30,362 frames}
& \textbf{Thermal weak-small UAV search}
& \textbf{\abyes}
& \textbf{\abyes}
& \textbf{\abyes}
& \textbf{\abyes} \\

\bottomrule
\end{tabular}
\end{adjustbox}
\vspace{-1mm}
\end{table*}

Table~\ref{tab:benchmark_core_advantages} positions TIR-UAV120-Gaze relative to representative gaze, saliency, and Anti-UAV datasets.
The compared resources usually provide either human-attention annotations or thermal weak-small-target data.
In contrast, TIR-UAV120-Gaze combines thermal infrared scenes, real eye tracking, task-driven target search, and weak-small UAV detection in one benchmark.
This combination provides the paired data required for both real-gaze supervision and the gaze--box relation modeling used in pseudo-gaze transfer.

\subsubsection{Data Collection and Annotation Protocol}
\label{sec:benchmark_annotations}

\begin{figure*}[t]
    \centering
    \includegraphics[
        width=\textwidth,
        trim=0.5cm 0cm -0.5cm 0cm,
        clip
    ]{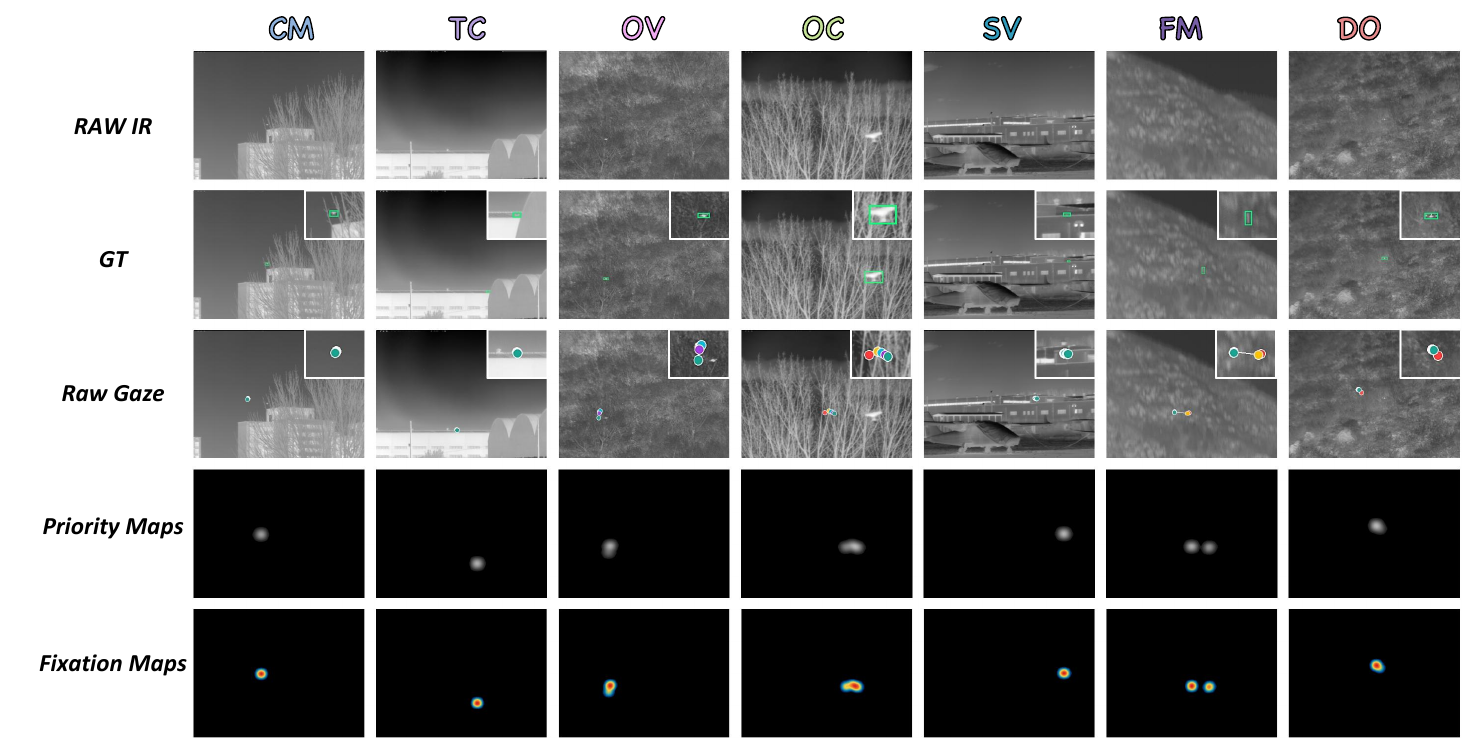}
    \caption{Representative TIR-UAV120-Gaze samples under seven challenge attributes:
    camera motion (CM), thermal crossover (TC), out-of-view (OV), occlusion (OC), scale variation (SV), fast motion (FM), and distractor objects (DO).
    The rows show the raw infrared frame, ground-truth bounding box, raw gaze samples, gaze-coordinate priority-supervision map, and fixation-density map.
    Insets enlarge the target or gaze region for visualization.
    The eye-tracking annotations are used only to construct training-time priority-supervision maps and are not required during inference.}
    \label{fig:benchmark_gaze_priority_examples}
\end{figure*}

The eye-tracking task was designed to match the target-search objective of thermal weak-small UAV detection.
During data collection, observers viewed thermal infrared UAV videos and searched for the UAV target.
Their eye movements were recorded using an EyeLink 1000 eye tracker.
The recorded gaze stream was then temporally aligned with the sampled video frames before frame-level gaze annotations were generated.
We have obtained the approvals of all the human participants.
Every participant has signed an informed consent form before the experiment.

The detection annotations follow the standard object-detection setting.
A target-present frame contains an annotated UAV bounding box, whereas a target-absent frame contains no visible UAV box.
Target-absent frames are retained because they are necessary for measuring false-positive detections in background-only observations.
However, they are not treated as positive real-gaze priority-supervision samples in the main protocol, because gaze in a target-absent frame does not define the location of a visible UAV target.

After temporal alignment, each sampled frame is associated with two gaze representations.
First, the frame-level gaze descriptor records the gaze location together with its validity and strength information.
Second, the fixation-density map aggregates the aligned gaze observations into a dense spatial distribution.
The gaze descriptor preserves a compact representation of frame-level gaze behavior, whereas the fixation-density map describes where attention is spatially concentrated during target search.

The gaze annotations and detection annotations provide different but complementary information.
A bounding box specifies the final location and spatial extent of the UAV.
In contrast, a fixation-density map describes the spatial distribution of task-driven search and may cover the target, its surrounding region, or multiple target-relevant locations.
Therefore, fixation-density maps are used to construct real-gaze priority-supervision maps rather than to replace bounding-box annotations.
The conversion from fixation-density maps to normalized real-gaze priority-supervision maps is described in Section~\ref{sec:gaze_target_construction}.

Figure~\ref{fig:benchmark_gaze_priority_examples} shows that gaze distributions do not necessarily coincide exactly with the target bounding boxes.
For clear targets, gaze may concentrate near the target center.
Under occlusion, fast motion, or target disappearance, the gaze distribution may instead extend to the surrounding search region.
These examples illustrate the intended role of gaze supervision: it describes task-related spatial preference before final localization, while the bounding box continues to define the final detection target.

\subsubsection{Dataset Split and Statistics}
\label{sec:benchmark_statistics}

To avoid information leakage between temporally adjacent frames, TIR-UAV120-Gaze is divided at the video level.
The resulting split is trial-disjoint, meaning that all frames from one video remain in the same partition.
Specifically, 90 videos with 20,635 frames are used for training, while the remaining 30 videos with 9,727 frames are used for testing.
This sequence-level separation prevents visually similar neighboring frames from appearing in both training and test sets.

\begin{table*}[t]
\centering
\PaperTableFont
\setlength{\tabcolsep}{3.0pt}
\renewcommand{\arraystretch}{1.05}
\caption{Split-level detection and gaze-file statistics of TIR-UAV120-Gaze.
Gaze descriptors and fixation-density files are provided for every sampled frame; invalid gaze and target-absent frames are excluded from positive real-gaze priority supervision.}
\label{tab:benchmark_split_stats}
\begin{tabular*}{\textwidth}{@{\extracolsep{\fill}}lcccccccc@{}}
\toprule
\textbf{Split} &
\textbf{Videos} &
\textbf{Frames} &
\textbf{Image Size} &
\textbf{Present} &
\textbf{Absent} &
\textbf{Present Rate} &
\textbf{Gaze Files} &
\textbf{Fix. Map Files} \\
\midrule
Train
& 90
& 20,635
& $640{\times}480$
& 20,027
& 608
& 97.05\%
& 20,635
& 20,635 \\

Test
& 30
& 9,727
& $640{\times}480$
& 9,549
& 178
& 98.17\%
& 9,727
& 9,727 \\

Total
& 120
& 30,362
& $640{\times}480$
& 29,576
& 786
& 97.41\%
& 30,362
& 30,362 \\

\bottomrule
\end{tabular*}
\end{table*}

Table~\ref{tab:benchmark_split_stats} summarizes the split-level statistics.
Across the complete benchmark, 29,576 frames contain an annotated UAV, while 786 frames are target-absent.
Thus, target-present frames account for 97.41\% of all sampled frames.
The test set contains 9,549 target-present frames and 178 target-absent frames.
Among the 9,549 target-present test frames, 9,452 have valid frame-aligned gaze records.
These test-set gaze records are used only for dataset characterization and are never provided to the detector or used to compute detection metrics.
Frames with invalid gaze records remain in the detection evaluation.

The benchmark also provides challenge and scene annotations for analyzing detection behavior under different conditions.
The seven challenge attributes are camera motion, thermal crossover, out-of-view, occlusion, scale variation, fast motion, and distractor objects.
The four scene labels are building, trees, mountain, and sky.
Because both challenge and scene annotations are multi-label, one frame may belong to more than one category.
Therefore, the occurrence percentages of these labels are computed independently and are not expected to sum to 100\%.

\begin{figure*}[t]
    \centering
    \includegraphics[
        width=0.95\textwidth,
        trim={0pt 0pt 0pt 0pt},
        clip
    ]{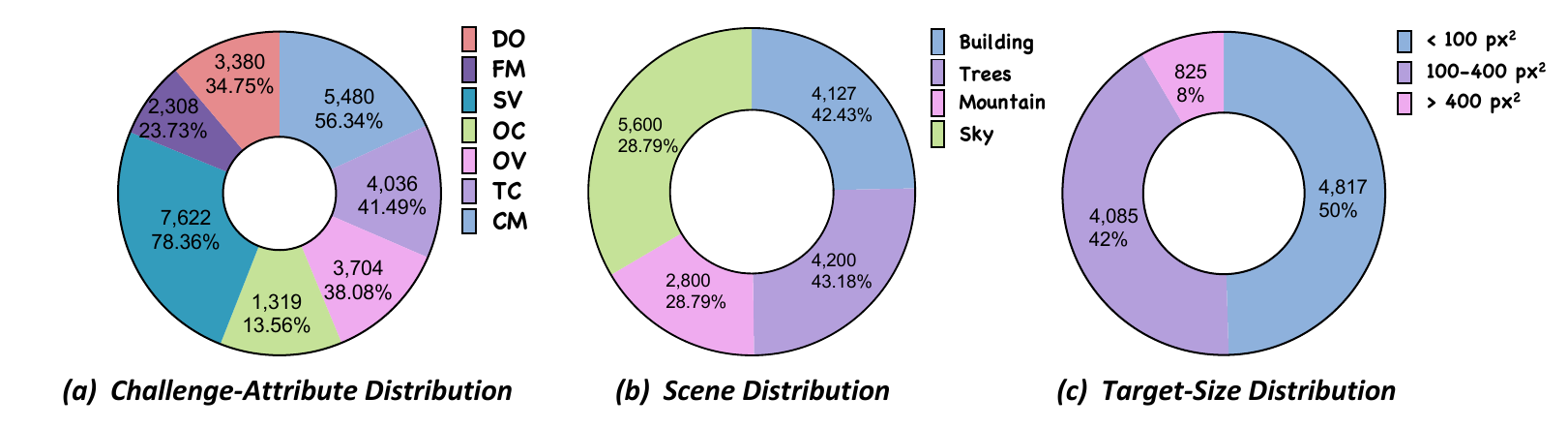}
    \caption{Test-set statistics of TIR-UAV120-Gaze.
    (a) Occurrence frequency of the seven multi-label challenge attributes.
    (b) Occurrence frequency of the four multi-label scene categories.
    Percentages in (a) and (b) are computed relative to the 9,727 test frames and therefore need not sum to 100\%.
    (c) Target-size distribution over the 9,549 target-present test frames, grouped by bounding-box area.}
    \label{fig:benchmark_distribution_stats}
\end{figure*}

Figure~\ref{fig:benchmark_distribution_stats} summarizes the challenge, scene, and target-size distributions of the test split.
The challenge statistics show that the test set includes both appearance-related difficulties, such as thermal crossover and occlusion, and motion-related difficulties, such as camera motion and fast motion.
Meanwhile, the scene annotations cover buildings, trees, mountains, and sky backgrounds, which provide different forms of structural clutter.

The target-size statistics further confirm the weak-small-target focus of the benchmark.
Among the 9,549 target-present test frames, 4,639 fall into the $<100$ px$^2$ bin, 4,085 fall into the $100$--$400$ px$^2$ bin, and 825 fall into the $>400$ px$^2$ bin.
Accordingly, 91.36\% of the target-present test frames contain targets smaller than 400 px$^2$.
Consequently, the benchmark is dominated by weak-small targets rather than medium or large objects.

Taken together, the trial-disjoint split, target-absent frames, multi-label challenge annotations, diverse scene categories, and paired eye-tracking annotations support several complementary evaluations.
The detection labels enable standard localization and false-positive analysis.
The challenge and scene labels enable condition-specific diagnostic evaluation.
Finally, the gaze annotations enable controlled study of real-gaze priority supervision and support the gaze--box relation modeling used for pseudo-gaze transfer.

\section{Experiments}
\label{sec:experiments}

This section evaluates Gaze-DETR on TIR-UAV120-Gaze and Anti-UAV410.
We first describe the datasets, evaluation protocol, and training details.
We then compare the proposed supervision schemes with representative methods and analyze their precision--recall and qualitative detection behavior.
Next, module and attribute-level ablations identify the sources and conditions of the performance gains, while priority-head visualizations examine the learned spatial representation.
Finally, cross-detector experiments and the discussion assess the generality and practical limitations of the proposed formulation.

\subsection{Datasets}
\label{sec:implementation_details}
Following the benchmark construction in Section~\ref{sec:benchmark_construction}, we train on 90 sequences with 20,635 frames and evaluate on 30 unseen sequences with 9,727 frames under the trial-disjoint protocol.
Real task-driven gaze is used only to construct training-time real-gaze priority-supervision maps; no gaze annotation from the test split is provided to the detector.

\begin{table*}[t!]
\centering
\PaperTableFont

\def\SOTAColSep{1.8pt}    % normal column padding
\def\SOTAGroupSep{0.8pt}  % padding around vertical group rules
\setlength{\tabcolsep}{\SOTAColSep}
\renewcommand{\arraystretch}{1.05}

\caption{Unified comparison on TIR-UAV120-Gaze and Anti-UAV410. The proposed rows use real-gaze supervision on TIR-UAV120-Gaze, transferred pseudo-gaze supervision on Anti-UAV410, or low-cost box-derived supervision. Red and blue denote the best and second-best results in each metric column, respectively.}
\label{tab:quantitative_uav121_antiuav410_combined}

 \begin{adjustbox}{width=\textwidth}
\begin{tabular}{
l
!{\hspace{\SOTAGroupSep}\vrule\hspace{\SOTAGroupSep}}
c
!{\hspace{\SOTAGroupSep}\vrule\hspace{\SOTAGroupSep}}
c
!{\hspace{\SOTAGroupSep}\vrule\hspace{\SOTAGroupSep}}
cccc
!{\hspace{\SOTAGroupSep}\vrule\hspace{\SOTAGroupSep}}
cccc
!{\hspace{\SOTAGroupSep}\vrule\hspace{\SOTAGroupSep}}
ccc
}
\toprule
\multirow{2}{*}{\textbf{Method}} & 
\multirow{2}{*}{\textbf{Publication}} &
\multirow{2}{*}{\textbf{Input Size}} &
\multicolumn{4}{c!{\hspace{\SOTAGroupSep}\vrule\hspace{\SOTAGroupSep}}}{\textbf{TIR-UAV120-Gaze}} & 
\multicolumn{4}{c!{\hspace{\SOTAGroupSep}\vrule\hspace{\SOTAGroupSep}}}{\textbf{Anti-UAV410}} & 
\multicolumn{3}{c}{\textbf{Complexity / Speed}} \\
\cmidrule(lr){4-7}\cmidrule(lr){8-11}\cmidrule(lr){12-14}
& & &
\textbf{mAP$_{50}\uparrow$} & 
\textbf{P$\uparrow$} & 
\textbf{R$\uparrow$} & 
\textbf{F1$\uparrow$} & 
\textbf{mAP$_{50}\uparrow$} & 
\textbf{P$\uparrow$} & 
\textbf{R$\uparrow$} & 
\textbf{F1$\uparrow$} & 
\textbf{GFLOPs$\downarrow$} & 
\textbf{Params (M)$\downarrow$} & 
\textbf{FPS$\uparrow$} \\
\midrule

Deformable-DETR~\cite{zhu2021deformable} & ICLR'21 & $512{\times}512$
& 80.47 & \textcolor{red}{\textbf{96.24}} & 81.59 & 88.31 
& 81.43 & 93.80 & 82.32 & 87.69 
& 143.42 & 47.32 & 28.99 \\

YOLOv6n~\cite{li2022yolov6} & ArXiv'22 & $512{\times}512$
& 79.05 & 88.10 & 69.90 & 78.00 
& 85.47 & 92.70 & 79.30 & 85.50 
& 7.88 & 5.04 & 104.43 \\

YOLOv7~\cite{wang2022yolov7} & ArXiv'22 & $512{\times}512$
& 70.33 & 85.69 & 57.55 & 68.86 
& 86.70 & 96.40 & 79.80 & 87.32 
& 8.33 & 6.01 & \textcolor{red}{\textbf{198.33}} \\

DINO-DETR~\cite{zhang2023dino} & ICLR'23 & $512{\times}512$
& 84.34 & 88.19 & \textcolor{red}{\textbf{85.65}} & 86.90 
& 83.37 & 93.72 & 79.00 & 85.73 
& 162.48 & 46.82 & 25.80 \\

YOLOv8n~\cite{jocher2023yolov8} & Ultralytics'23 & $512{\times}512$
& 77.18 & 92.67 & 70.70 & 80.21 
& 85.80 & 94.90 & 80.00 & 86.81 
& 5.17 & 3.01 & 160.04 \\

RT-DETR-l~\cite{zhao2024rtdetr} & CVPR'24 & $512{\times}512$
& 79.34 & 89.75 & 80.20 & 84.70 
& 86.42 & 91.35 & 83.60 & 87.31 
& 67.43 & 31.99 & 24.13 \\

YOLOv9t~\cite{wang2024yolov9} & ArXiv'24 & $512{\times}512$
& 78.19 & 92.51 & 72.01 & 80.98 
& 86.03 & 95.11 & 79.60 & 86.67 
& 4.86 & \textcolor{red}{\textbf{2.01}} & 68.73 \\

YOLOv10n~\cite{wang2024yolov10} & NeurIPS'24 & $512{\times}512$
& 77.81 & 89.63 & 71.24 & 79.38 
& 79.16 & 90.96 & 67.24 & 77.32 
& 5.27 & 2.69 & \textcolor{blue}{\textbf{165.99}} \\

YOLO11n~\cite{ultralytics2024yolo11} & Ultralytics'24 & $512{\times}512$
& 79.40 & 92.58 & 72.89 & 81.57 
& 84.74 & 93.51 & 77.47 & 84.74 
& 4.04 & 2.59 & 142.84 \\

SSTNet~\cite{chen2024sstnet} & TGRS'24 & $512{\times}512$
& 76.95 & 88.67 & 74.63 & 81.05
& 86.19 & 93.63 & 81.87 & 87.35
& 123.59 & 11.95 & 10.61 \\

YOLO12n~\cite{tian2025yolov12} & ArXiv'25 & $512{\times}512$
& 78.75 & 91.19 & 73.79 & 81.57 
& 84.79 & 95.09 & 77.99 & 85.70 
& 4.15 & 2.57 & 84.75 \\

YOLOv13n~\cite{lei2025yolov13} & ArXiv'25 & $512{\times}512$
& 74.08 & 87.51 & 64.93 & 74.55 
& 85.18 & 94.94 & 77.60 & 85.40 
& \textcolor{blue}{\textbf{3.97}} & \textcolor{blue}{\textbf{2.45}} & 73.6 \\

PConv~\cite{yang2025pconv} & AAAI'25 & $512{\times}512$
& 67.50 & 94.07 & 72.37 & 81.81 
& 85.67 & 97.70& 73.90 & 84.15 
& 7.89 & 2.91 & 44.92 \\

L2SKNet~\cite{wu2025l2sknet} & TGRS'25 & $512{\times}512$
& 76.78 & 86.74 & 72.12 & 78.76
& 81.66 & \textcolor{red}{\textbf{98.77}}  & 64.71 & 78.19
& 76.00 & 3.42 & 40.64 \\

YOLO26n~\cite{ultralytics2026yolo26} & Ultralytics'26 & $512{\times}512$
& 78.56 & 92.97 & 72.04 & 81.18 
& 86.49 & 95.13 & 79.28 & 86.49 
& \textcolor{red}{\textbf{3.32}} & 2.50 & 123.66 \\

CHAL~\cite{duan2026chal} & CVPR'26 & $512{\times}512$
& 72.50 & 83.72 & 68.23 & 75.19 
& 80.95 & 93.34 & 74.63 & 82.9 
& 167.04 & 15.69 & 2.42 \\

\rowcolor{gray!15}
\textbf{Gaze-DETR (gaze/pseudo)} & -- & \textbf{$512{\times}512$}
& \textcolor{red}{\textbf{86.18}} & \textcolor{blue}{\textbf{94.64}} & \textcolor{blue}{\textbf{84.00}} & \textcolor{red}{\textbf{89.00}} 
& \textcolor{red}{\textbf{87.08}} & \textcolor{blue}{\textbf{98.32}} & \textcolor{blue}{\textbf{83.71}} & \textcolor{blue}{\textbf{90.43}} 
& 167.52 & 47.43 & 21.01 \\

\rowcolor{gray!15}
\textbf{Gaze-DETR (box-derived)} & -- & \textbf{$512{\times}512$}
& \textcolor{blue}{\textbf{85.76}} & 94.12 & \textcolor{blue}{\textbf{84.00}} & \textcolor{blue}{\textbf{88.77}}
& \textcolor{blue}{\textbf{87.06}} & 96.38 & \textcolor{red}{\textbf{86.00}} & \textcolor{red}{\textbf{90.90}} 
& 167.32 & 47.42 & 21.48 \\

\bottomrule
\end{tabular}
\end{adjustbox}
\vspace{-2mm}
\end{table*}

\begin{figure*}[t]
    \centering
    \includegraphics[
        width=\textwidth,
        trim=0cm 0cm -0.5cm 0cm,
        clip
    ]{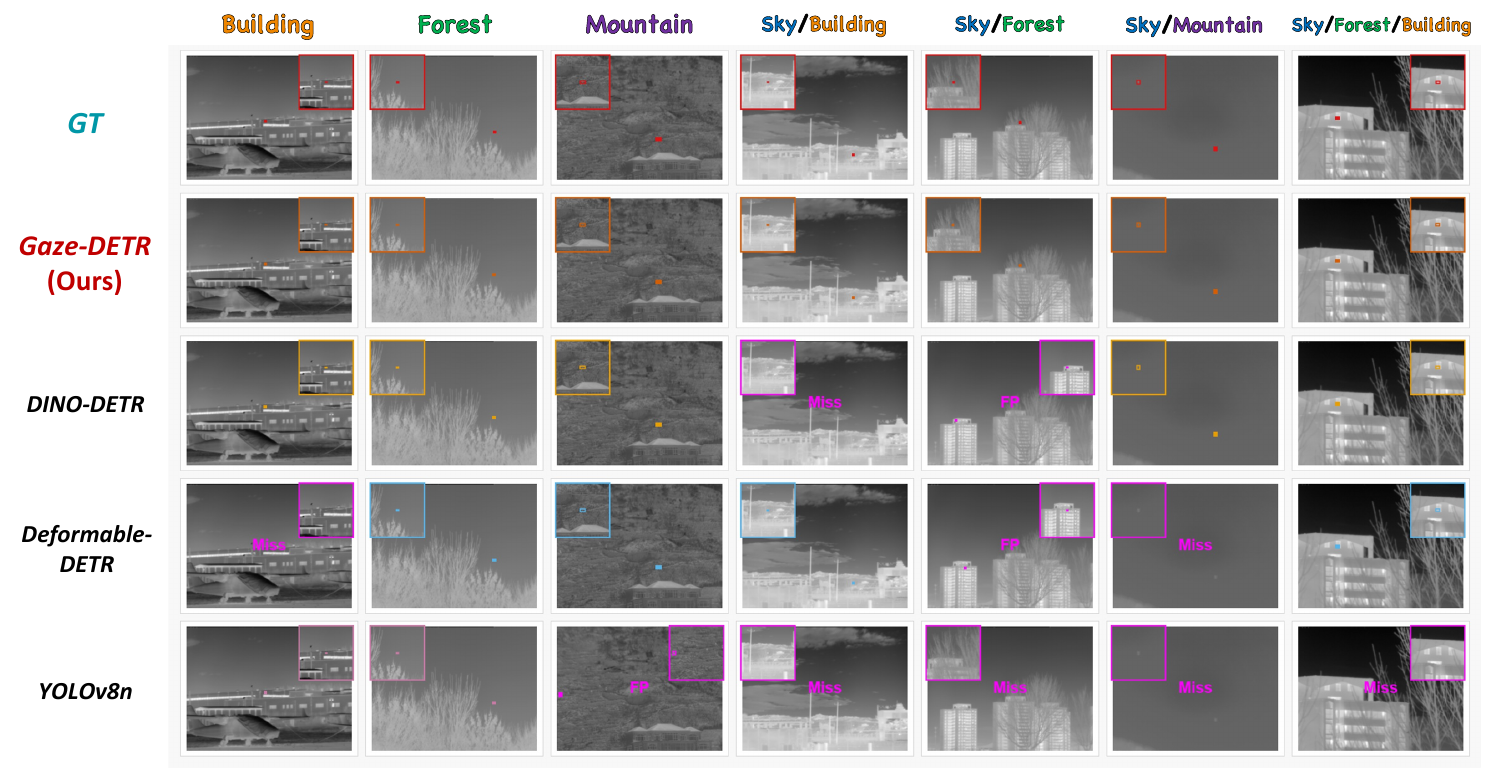}
    \caption{Qualitative comparison on challenging TIR-UAV120-Gaze scenes using the real-gaze-supervised Gaze-DETR. Rows show ground truth, Gaze-DETR, and reproduced baselines. Magenta labels mark representative missed detections or false positives.}
    \vspace{-3mm}
    \label{fig:qualitative_sota_comparison}
\end{figure*}

Anti-UAV410 is a public thermal UAV benchmark without eye-tracking annotations~\cite{huang2024antiuav410}.
We follow its official split of 200 training, 90 validation, and 120 test sequences.
After conversion to the unified COCO format, the three splits contain 213,995, 129,691, and 94,711 frames, respectively.
Target-absent frames are retained on both benchmarks so that false-positive predictions are included in the evaluation.

\subsection{Metrics and Evaluation Protocol}
We report mAP$_{50}$, precision (P), recall (R), and F1.
The operating-point metrics are defined as
\begin{equation}
\mathrm{P}=\frac{\mathrm{TP}}{\mathrm{TP}+\mathrm{FP}},\quad
\mathrm{R}=\frac{\mathrm{TP}}{\mathrm{TP}+\mathrm{FN}},\quad
\mathrm{F1}=\frac{2\mathrm{P}\mathrm{R}}{\mathrm{P}+\mathrm{R}}.
\end{equation}
During testing, only predicted boxes with confidence scores greater than 0.001 are retained.
The IoU threshold for matching predicted boxes with ground-truth boxes is set to 0.50.
Within each benchmark, all reproduced methods use the same annotations, $512\times512$ input resolution, IoU matching rule, and evaluation code.
DETR-based models use standard COCO post-processing without nonmaximum suppression, whereas other detectors retain their native post-processing.
GFLOPs and parameter counts are calculated at $512\times512$ resolution, and FPS is measured with batch size 1 on a single NVIDIA A100 GPU under the same environment.

\begin{figure*}[t]
    \centering
    \includegraphics[
        width=0.98\textwidth,
        trim=0.25cm 0cm -0.5cm 0cm,
        clip
    ]{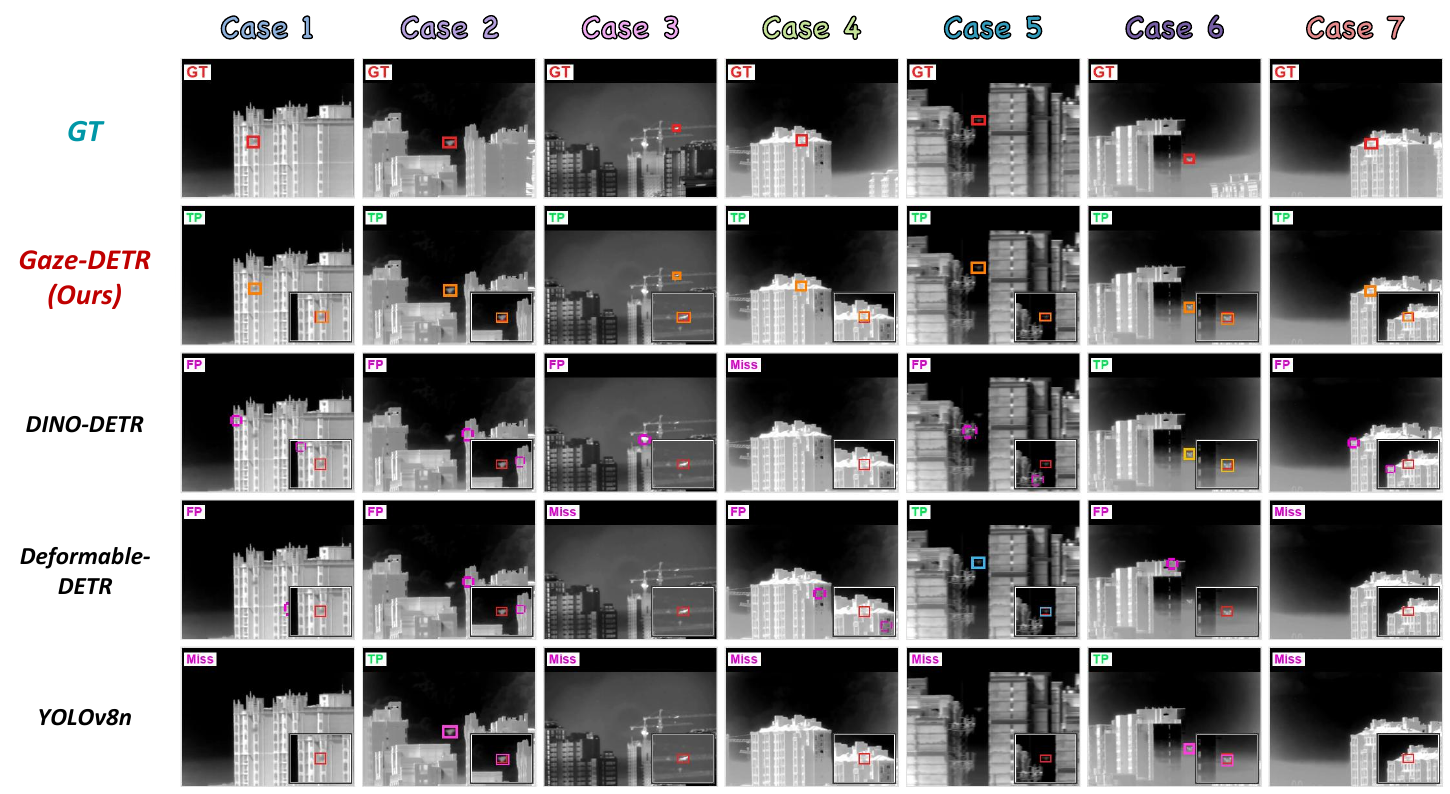}
    \caption{Qualitative comparison on Anti-UAV410 using the pseudo-gaze-supervised Gaze-DETR. Each column shows one infrared frame, with ground truth followed by Gaze-DETR, DINO-DETR, Deformable-DETR, and YOLOv8n. TP, FP, and Miss denote true positives, false positives, and missed detections.}
    \label{fig:uav410_teaser}
\end{figure*}

% \begin{figure*}[t]
%     \centering
%     \includegraphics[
%         width=\textwidth,
%         trim=0cm 0cm -0.5cm 0cm,
%         clip
%     ]{images/sota-vilualize-uav120.pdf}
%     \caption{Qualitative comparison on challenging TIR-UAV120-Gaze scenes using the real-gaze-supervised Gaze-DETR. Rows show ground truth, Gaze-DETR, and reproduced baselines. Magenta labels mark representative missed detections or false positives.}
%     \vspace{-3mm}
%     \label{fig:qualitative_sota_comparison}
% \end{figure*}

% \begin{figure*}[t]
%     \centering
%     \includegraphics[
%         width=0.98\textwidth,
%         trim=0.25cm 0cm -0.5cm 0cm,
%         clip
%     ]{images/uav410_multimodel_miss_fp_visual_labels_clean_conf070_iou050_20260706.pdf}
%     \caption{Qualitative comparison on Anti-UAV410 using the pseudo-gaze-supervised Gaze-DETR. Each column shows one infrared frame, with ground truth followed by Gaze-DETR, DINO-DETR, Deformable-DETR, and YOLOv8n. TP, FP, and Miss denote true positives, false positives, and missed detections.}
%     \label{fig:uav410_teaser}
% \end{figure*}

\subsection{Training Details}
% The reference implementation follows the completed TIR-UAV120-Gaze box-derived M4 configuration.
% It uses a four-scale DINO-DETR with a ResNet-50 backbone, 900 original detector queries, a $64\times64$ priority head, RPFM at all four feature levels, and gated PAQI with at most six additional queries.
% During training, the image short side is randomly sampled from 480 to 800 pixels under the standard DINO multi-scale resize and crop-resize augmentation, while the long side is capped at 1333 pixels.
% Horizontal flipping is also applied.
% At test time, the aspect ratio is preserved, the short side is set to 800 pixels, and the long side remains capped at 1333 pixels.

The main configuration uses a four-scale DINO-DETR with a ResNet-50 backbone, 900 original detector queries, and a $64\times64$ priority head; RPFM is enabled at all four feature levels.
We set $\lambda_p=0.5$ and $\lambda_{sim}=1$.
PAQI is activated after a four-epoch warm-up and adds at most six priority-guided queries from two spatially separated peaks and three training-set size priors, with a suppression radius of two priority-grid cells.
We train for 12 epochs on NVIDIA A100 40-GB GPUs with a global batch size of 40 and random seed 42.
AdamW uses learning rates of $1\times10^{-5}$ for non-backbone parameters and $1\times10^{-6}$ for the backbone, with weight decay $1\times10^{-4}$ and gradient clipping at 0.1; the learning rate is reduced by a factor of 0.1 after epoch 9.
An exponential moving average with decay 0.9997 is maintained throughout training.
The same optimization schedule is used for DINO-DETR and all priority-supervised variants.

% The priority objective combines KL divergence and histogram-intersection similarity with unit internal weights and is scaled by $\lambda_p=0.5$ in the total loss.
% RPFM is enabled at every feature level, with each residual modulation coefficient initialized to zero.
% PAQI retains all 900 original DINO queries and adds six priority-guided queries formed from two spatially separated priority peaks and three weak-target size priors.
% Peak suppression uses a radius of 2.
% PAQI is enabled after a four-epoch warm-up and accepts a predicted priority map only when its maximum peak is at least 0.005, its normalized entropy is at most 0.85, and its top-1/top-2 peak ratio is at least 1.05.
% Proposal-score bias, ground-truth anchor oracles, and gaze-feature sampling are disabled.
% The box-derived M4 reference constructs Gaussian priority targets exclusively from training-set bounding boxes; the real-gaze and transferred pseudo-gaze variants change only the source of the training-time priority target.
% No external priority map, gaze annotation, or test-set box is read during inference.

% For the present manuscript version, the current Gaze-DETR numerical entries are retained as final reporting values from their completed checkpoint evaluations.
% The box-derived M4 configuration above defines the standard implementation.
% A matched full-stack rerun replaces a retained value only after completing the fixed 12-epoch schedule, saving its final regular and EMA weights, passing the RPFM/PAQI parameter-activity audit, and producing one final image-only held-out evaluation.

\subsection{Comparison with State-of-the-Art Methods}
\label{sec:compare_sota_two_benchmarks}

Table~\ref{tab:quantitative_uav121_antiuav410_combined} compares Gaze-DETR with representative DETR, YOLO, and ISTD detectors under the unified protocol.
The two Gaze-DETR variants use the same architecture and differ only in the source of priority supervision.
On TIR-UAV120-Gaze, the real-gaze scheme constructs priority-supervision maps from fixation-density maps, whereas the box-derived scheme constructs Gaussian priority-supervision maps from training boxes.
On Anti-UAV410, real gaze is unavailable; therefore, the transferred pseudo-gaze scheme uses transferred pseudo-gaze priority-supervision maps.
This design isolates the effect of the supervision source while preserving image-only inference for all reported models.

On TIR-UAV120-Gaze, DINO-DETR obtains 84.34 mAP$_{50}$ and 86.90 F1.
Box-derived priority supervision increases these values to 85.76 and 88.77, while real-gaze supervision further reaches 86.18 mAP$_{50}$ and 89.00 F1.
Both schemes raise precision from 88.19 to more than 94, although recall decreases from 85.65 to 84.00.
Therefore, the F1 improvement is associated mainly with a more precise operating point rather than a recall increase.
Taken together, both settings improve the baseline, while their small performance difference should be interpreted together with their different annotation requirements.

On Anti-UAV410, DINO-DETR obtains 83.37 mAP$_{50}$ and 85.73 F1.
The box-derived scheme reaches 87.06 mAP$_{50}$ and 90.90 F1, whereas the pseudo-gaze scheme reaches 87.08 mAP$_{50}$ and 90.43 F1.
Although pseudo-gaze gives the highest mAP$_{50}$ and precision, box-derived supervision gives higher recall and F1.
Thus, neither supervision source is uniformly superior across metrics.
Instead, both results support the central claim that an explicit priority objective can complement the original detection losses under different annotation budgets.

The absolute results across the two benchmarks should also be interpreted together with their different target-scale distributions.
TIR-UAV120-Gaze is concentrated in an extremely low-area regime: 48.58\% of its target-present test frames contain boxes smaller than 100 px$^2$, and 91.36\% contain boxes smaller than 400 px$^2$.
Anti-UAV410 instead defines scale by bounding-box diagonal length on $640\times512$ frames; its official dataset-level statistics divide targets into tiny $[2,10)$, small $[10,30)$, medium $[30,50)$, and normal $[50,\infty)$ intervals, with 55.12\% of annotated targets below a 50-pixel diagonal and 44.88\% at or above 50 pixels~\cite{huang2024antiuav410}.
Because area and diagonal length are different scale measures, these distributions are not directly interchangeable.
Nevertheless, TIR-UAV120-Gaze places most test targets in a very low-pixel regime, whereas Anti-UAV410 covers a broader scale range and provides substantially more training frames.
This difference can contribute to the slightly higher absolute Gaze-DETR scores on Anti-UAV410, while that benchmark remains challenging because of dynamic background clutter, thermal crossover, and scale variation.
Thus, the cross-dataset performance gap reflects target scale, training-set size, and scene difficulty jointly rather than the supervision source alone.

The improvement introduces a moderate computational overhead.
Relative to DINO-DETR, the two Gaze-DETR variants add approximately 4.84--5.04 GFLOPs and 0.60--0.61M parameters, corresponding to about 3.0--3.1\% more computation and 1.3\% more parameters.
Their inference speed decreases from 25.80 FPS to 21.01--21.48 FPS because priority prediction, multi-scale modulation, and additional anchor queries are executed at inference.

\subsection{Precision--Recall and Qualitative Detection Behavior}
\label{sec:detection_behavior}

% \begin{figure*}[t!]
%     \centering
%     \includegraphics[width=0.98\textwidth]{images/pr_curve_two_datasets_gazepseudo_inset_bold_nozoom_20260703.pdf}\par\vspace{-1.0mm}
%     {\footnotesize (a) Real-gaze scheme on TIR-UAV120-Gaze and transferred pseudo-gaze scheme on Anti-UAV410.}\par\vspace{1.5mm}
%     \includegraphics[width=0.98\textwidth]{images/pr_curve_two_datasets_boxderived_inset_bold_nozoom_20260703.pdf}\par\vspace{-1.0mm}
%     {\footnotesize (b) Low-cost box-derived scheme on TIR-UAV120-Gaze and Anti-UAV410.}
%     \caption{Precision--recall curves for the evaluated priority-supervision settings. Panel (a) reports real-gaze supervision on TIR-UAV120-Gaze and transferred pseudo-gaze supervision on Anti-UAV410; panel (b) reports box-derived supervision on both benchmarks. All models are image-only at inference.}
%     \label{fig:pr_curve_two_benchmarks}
% \end{figure*}

\begin{figure*}[t!]
    \centering
    \includegraphics[width=0.98\textwidth]{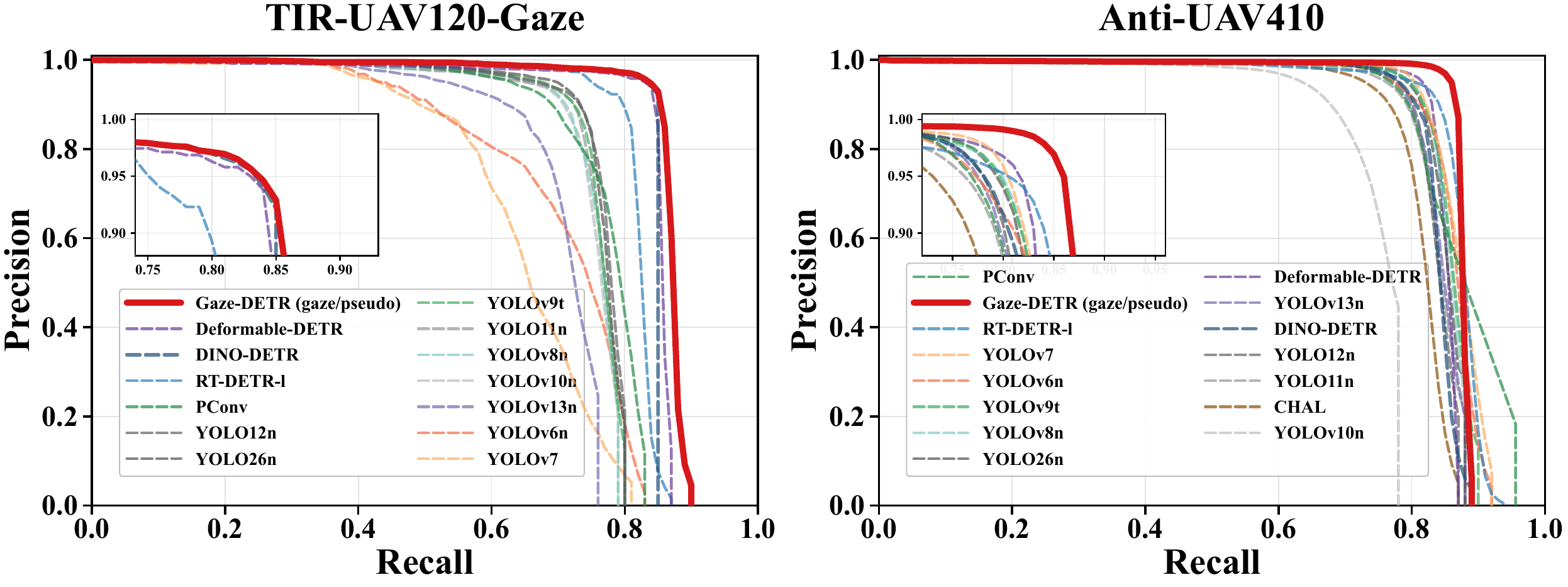}\par\vspace{-1.0mm}
    {\footnotesize (a) Real-gaze scheme on TIR-UAV120-Gaze and transferred pseudo-gaze scheme on Anti-UAV410.}\par\vspace{1.5mm}
    \includegraphics[width=0.98\textwidth]{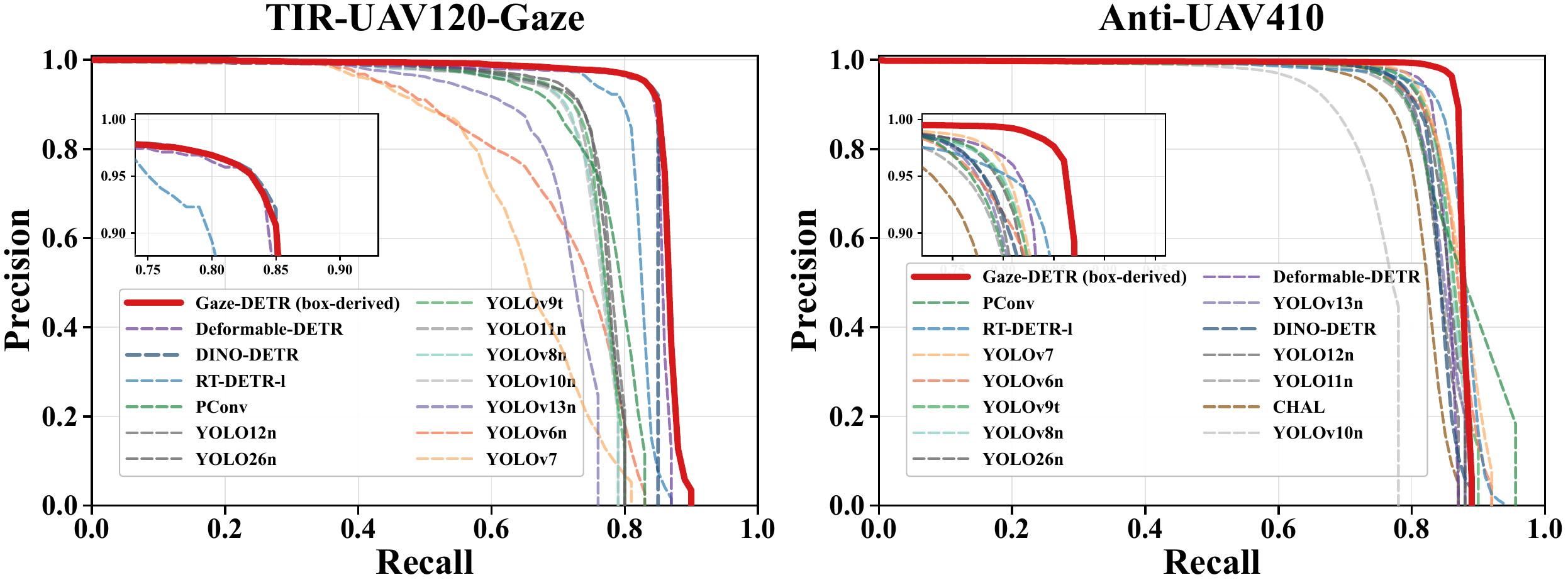}\par\vspace{-1.0mm}
    {\footnotesize (b) Low-cost box-derived scheme on TIR-UAV120-Gaze and Anti-UAV410.}
    \caption{Precision--recall curves for the evaluated priority-supervision settings. Panel (a) reports real-gaze supervision on TIR-UAV120-Gaze and transferred pseudo-gaze supervision on Anti-UAV410; panel (b) reports box-derived supervision on both benchmarks. All models are image-only at inference.}
    \label{fig:pr_curve_two_benchmarks}
\end{figure*}

Figure~\ref{fig:pr_curve_two_benchmarks} compares the supervision schemes over the complete precision--recall range.
Panel (a) reports real-gaze supervision on TIR-UAV120-Gaze and pseudo-gaze supervision on Anti-UAV410, whereas panel (b) reports box-derived supervision on both benchmarks.
The Gaze-DETR curves remain competitive across a broad range of recall values, showing that the improvements in Table~\ref{tab:quantitative_uav121_antiuav410_combined} are not produced by a single confidence threshold.
At the same time, the different curve shapes agree with the operating-point results: the two supervision sources produce different precision--recall trade-offs rather than a fixed ranking.

% \begin{figure*}[t]
%     \centering
%     \includegraphics[
%         width=\textwidth,
%         trim=0cm 0cm -0.5cm 0cm,
%         clip
%     ]{images/sota-vilualize-uav120.pdf}
%     \caption{Qualitative comparison on challenging TIR-UAV120-Gaze scenes using the real-gaze-supervised Gaze-DETR. Rows show ground truth, Gaze-DETR, and reproduced baselines. Magenta labels mark representative missed detections or false positives.}
%     \vspace{-3mm}
%     \label{fig:qualitative_sota_comparison}
% \end{figure*}

% \begin{figure*}[t]
%     \centering
%     \includegraphics[
%         width=0.98\textwidth,
%         trim=0.25cm 0cm -0.5cm 0cm,
%         clip
%     ]{images/uav410_multimodel_miss_fp_visual_labels_clean_conf070_iou050_20260706.pdf}
%     \caption{Qualitative comparison on Anti-UAV410 using the pseudo-gaze-supervised Gaze-DETR. Each column shows one infrared frame, with ground truth followed by Gaze-DETR, DINO-DETR, Deformable-DETR, and YOLOv8n. TP, FP, and Miss denote true positives, false positives, and missed detections.}
%     \label{fig:uav410_teaser}
% \end{figure*}

Figures~\ref{fig:qualitative_sota_comparison} and~\ref{fig:uav410_teaser} provide complementary visual evidence on the two benchmarks.
The TIR-UAV120-Gaze examples contain buildings, vegetation, mountains, mixed scenes, and extremely small targets, while the Anti-UAV410 examples contain strong structural clutter and target-like bright regions.
In the selected cases, Gaze-DETR localizes the target without the representative missed detections or false positives produced by one or more reproduced baselines.
These examples are consistent with the precision and F1 improvements in the quantitative comparison.
However, they are intended to illustrate typical detection behavior rather than to establish that one method succeeds in every scene.

\subsection{Ablation Study}
\label{sec:ablation_study}

\subsubsection{Contributions of Priority Loss, RPFM, and PAQI}

Table~\ref{tab:module_ablation_full} separates the auxiliary supervision effect from the two mechanisms that consume the predicted map.
With real-gaze supervision, the priority loss alone improves DINO-DETR by 0.91 mAP$_{50}$ and 1.50 F1, indicating that the auxiliary objective provides useful task-related spatial supervision to the shared image features.
Adding RPFM increases the gains to 1.44 mAP$_{50}$ and 1.84 F1, whereas adding PAQI increases them to 1.58 and 1.97, respectively.
When both modules are enabled, the full model achieves the best result of 86.18 mAP$_{50}$ and 89.00 F1.

The box-derived scheme exhibits the same progressive pattern.
The priority loss alone provides gains of 0.61 mAP$_{50}$ and 0.99 F1; RPFM and PAQI then contribute additional improvements, and their combination reaches 85.76 mAP$_{50}$ and 88.77 F1.
Therefore, the performance increase cannot be attributed only to an additional loss or only to extra decoder queries.
Instead, the results support the intended division of labor: the priority loss learns the spatial representation, RPFM introduces it into multi-scale feature processing, and PAQI uses its high-priority locations for candidate initialization.

\subsubsection{Behaviour under Difficult Attributes}

\begin{table}[!t]
\centering
\setlength{\tabcolsep}{3.0pt}
\renewcommand{\arraystretch}{1.15}
\setlength{\aboverulesep}{0pt}
\setlength{\belowrulesep}{0pt}
\setlength{\extrarowheight}{0pt}
\caption{Attribute-level scheme comparison on TIR-UAV120-Gaze. 
Each subset compares the bounding-box-only baseline, the low-cost box-derived scheme, and the high-cost real-gaze scheme. 
}
\label{tab:gaze_heatmap_attr_ablation}
\begin{adjustbox}{max width=\columnwidth}
\begin{tabular}{llcccccc}
\toprule
\textbf{Attr.} &
\textbf{Variant} &
\textbf{mAP$_{50}$} &
\textbf{$\Delta$mAP$_{50}$ (pts)} &
\textbf{F1} &
\textbf{$\Delta$F1 (pts)} &
\textbf{Frm.} &
\textbf{Box} \\
\midrule

\multirow{3}{*}{\textbf{CM}}
& Baseline & 80.79 & -- & 88.81 & -- & \multirow{3}{*}{5480} & \multirow{3}{*}{5419} \\
& Box-derived & 80.84 & \attrgain{0.05} & 88.61 & \attrloss{0.20} &  &  \\
\rowcolor{gray!15}
& Real gaze & 80.77 & \attrloss{0.02} & 88.99 & \attrgain{0.18} &  &  \\
\midrule

\multirow{3}{*}{\textbf{TC}}
& Baseline & 72.25 & -- & 82.58 & -- & \multirow{3}{*}{4036} & \multirow{3}{*}{3886} \\
& Box-derived & 71.65 & \attrloss{0.60} & 82.71 & \attrgain{0.13} &  &  \\
\rowcolor{gray!15}
& Real gaze & 72.88 & \attrgain{0.63} & 82.92 & \attrgain{0.34} &  &  \\
\midrule

\multirow{3}{*}{\textbf{OV}}
& Baseline & 80.75 & -- & 88.67 & -- & \multirow{3}{*}{3704} & \multirow{3}{*}{3560} \\
& Box-derived & 80.69 & \attrloss{0.06} & 88.44 & \attrloss{0.23} &  &  \\
\rowcolor{gray!15}
& Real gaze & 81.60 & \attrgain{0.85} & 88.93 & \attrgain{0.26} &  &  \\
\midrule

\multirow{3}{*}{\textbf{OC}}
& Baseline & 69.15 & -- & 80.86 & -- & \multirow{3}{*}{1319} & \multirow{3}{*}{1226} \\
& Box-derived & 70.15 & \attrgain{1.00} & 81.55 & \attrgain{0.69} &  &  \\
\rowcolor{gray!15}
& Real gaze & 71.09 & \attrgain{1.94} & 81.57 & \attrgain{0.71} &  &  \\
\midrule

\multirow{3}{*}{\textbf{SV}}
& Baseline & 75.25 & -- & 85.02 & -- & \multirow{3}{*}{7622} & \multirow{3}{*}{7444} \\
& Box-derived & 75.32 & \attrgain{0.07} & 84.89 & \attrloss{0.13} &  &  \\
\rowcolor{gray!15}
& Real gaze & 75.94 & \attrgain{0.69} & 85.21 & \attrgain{0.19} &  &  \\
\midrule

\multirow{3}{*}{\textbf{FM}}
& Baseline & 81.86 & -- & 89.41 & -- & \multirow{3}{*}{2308} & \multirow{3}{*}{2199} \\
& Box-derived & 81.63 & \attrloss{0.23} & 89.17 & \attrloss{0.24} &  &  \\
\rowcolor{gray!15}
& Real gaze & 82.74 & \attrgain{0.88} & 89.94 & \attrgain{0.53} &  &  \\
\midrule

\multirow{3}{*}{\textbf{DO}}
& Baseline & 85.95 & -- & 92.25 & -- & \multirow{3}{*}{3380} & \multirow{3}{*}{3331} \\
& Box-derived & 85.96 & \attrgain{0.01} & 91.85 & \attrloss{0.40} &  &  \\
\rowcolor{gray!15}
& Real gaze & 85.95 & \attrzero & 92.29 & \attrgain{0.04} &  &  \\
\bottomrule
\end{tabular}
\end{adjustbox}
\vspace{-2mm}
\end{table}

Table~\ref{tab:gaze_heatmap_attr_ablation} evaluates the two supervision schemes on the seven multi-label challenge subsets of TIR-UAV120-Gaze.
The box-derived scheme improves mAP$_{50}$ under camera motion, occlusion, scale variation, and distractor objects, and it improves F1 under thermal crossover and occlusion.
However, small decreases appear on several other metric--attribute pairs.
Therefore, a box-centered Gaussian provides useful low-cost supervision, but its effect is not uniform across scene conditions.

Real-gaze supervision gives non-negative F1 changes on all seven subsets and improves mAP$_{50}$ under thermal crossover, out-of-view, occlusion, scale variation, and fast motion.
Its largest mAP$_{50}$ gain is 1.94 points under occlusion, where the visible target evidence is incomplete.
By contrast, the changes under camera motion and distractor objects are limited.
Taken together, the two supervision schemes exhibit different condition-dependent behaviors, and neither produces a uniform gain across all attribute--metric combinations.
Real-gaze supervision provides larger gains under occlusion and fast motion in the current evaluation, whereas box-derived supervision remains effective under several conditions without additional eye-tracking cost.

\subsection{Analysis of Priority-Head Predictions}
\label{sec:priority_map_visualization}

The preceding quantitative results demonstrate that priority supervision improves detection performance.
However, these results do not directly reveal what spatial information is learned by the priority head.
Therefore, we visualize the predicted priority maps of the complete Gaze-DETR under diverse challenging scenes.
For TIR-UAV120-Gaze, the analyzed model is trained with real-gaze priority-supervision maps, whereas the Anti-UAV410 model is trained with transferred pseudo-gaze priority-supervision maps.
In both settings, the predicted priority map is produced from the infrared image alone.
The ground-truth annotations are included only to indicate the target locations for visual analysis and are not provided to the detector during inference.

\begin{figure*}[t]
    \centering
    \includegraphics[width=\textwidth]{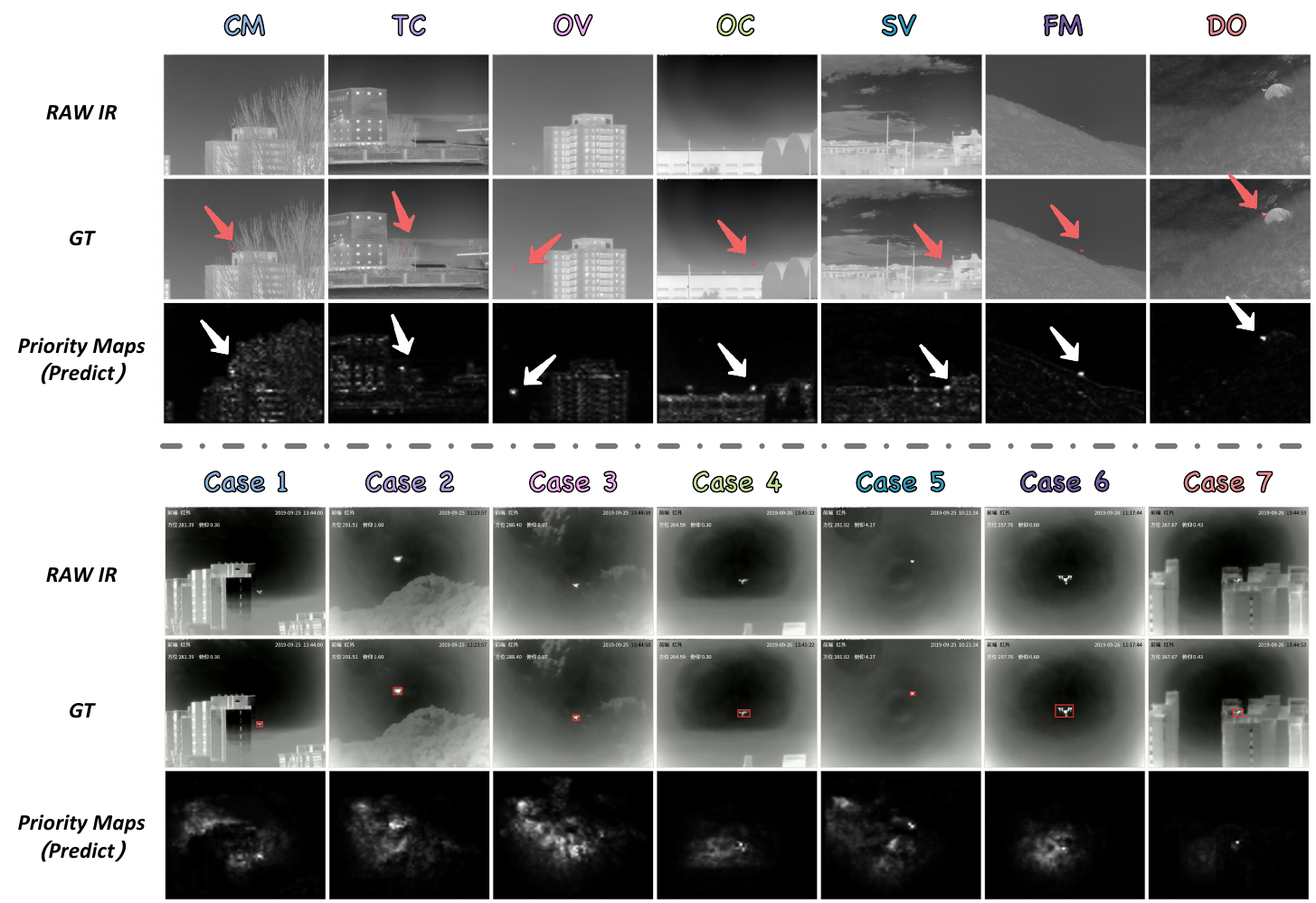}
    \caption{Visualization of priority-head predictions on TIR-UAV120-Gaze and Anti-UAV410.
    The upper panel shows results from the real-gaze-supervised Gaze-DETR on TIR-UAV120-Gaze.
    Its seven columns correspond to camera motion (CM), thermal crossover (TC), out-of-view (OV), occlusion (OC), scale variation (SV), fast motion (FM), and distractor objects (DO).
    The lower panel shows results from the pseudo-gaze-supervised Gaze-DETR on seven challenging Anti-UAV410 cases with different target appearances, scales, and background structures.
    For each dataset, the three rows show the raw infrared image, the ground-truth target location, and the predicted priority map produced from the infrared image alone.
    Brighter regions indicate higher relative spatial priority.
    Ground-truth annotations are displayed only for analysis and are not used during inference.}
    \label{fig:priority_prediction_combined}
\end{figure*}

As shown in the upper panel of Fig.~\ref{fig:priority_prediction_combined}, the priority head produces a target-related high-priority response under all seven selected TIR-UAV120-Gaze attributes.
Nevertheless, the spatial form of the predicted response varies with the scene.
For example, the responses under OV, FM, and DO are relatively concentrated around the annotated UAV locations.
In contrast, the maps under TC, OC, and SV contain broader responses that also cover nearby buildings, vegetation, or other structured regions.
Under CM, the priority head still preserves a target-related response despite camera-induced background variation.
These observations indicate that the predicted priority map is conditioned on the input scene rather than being a fixed Gaussian centered at a predetermined location.

The lower panel of Fig.~\ref{fig:priority_prediction_combined} further examines the behavior of the pseudo-gaze-supervised model on Anti-UAV410.
Across the seven selected cases, the predicted maps retain high-priority responses at or near the annotated UAV locations despite variations in target scale, thermal appearance, and background structure.
However, several maps are more spatially diffuse than those in the upper panel and assign secondary priority to buildings, clouds, vegetation, or other high-contrast structures.
This behavior may arise from both the greater background ambiguity of Anti-UAV410 and the difference between transferred pseudo-gaze supervision and real task-driven gaze.
Even so, the target-related regions remain represented among the high-priority candidates in the selected cases.

Taken together, the two groups of visualizations clarify the role of the priority head.
The predicted map is not a target segmentation mask and does not need to coincide exactly with the ground-truth bounding box.

\begin{table*}[t]
\centering
\setlength{\tabcolsep}{1.8pt}
\renewcommand{\arraystretch}{1.02}
\caption{Module-isolating ablation on TIR-UAV120-Gaze for the two priority-supervision schemes. The upper block reports the real-gaze scheme, and the lower block reports the box-derived scheme. The symbol ``--'' denotes a non-applicable or disabled component.}
\label{tab:module_ablation_full}
\begin{adjustbox}{max width=\textwidth}
\begin{tabular}{lcccc|cccc|c}
\toprule
\textbf{ID} & \textbf{Priority-supervision scheme} & \textbf{Priority loss} & \textbf{RPFM} & \textbf{PAQI} &
\textbf{mAP$_{50}$} & \textbf{P} & \textbf{R} & \textbf{F1} & \textbf{$\Delta$ mAP$_{50}$/F1} \\
\midrule
M0 & None & -- & -- & -- & 84.34 & 88.19 & 85.65 & 86.90 & -- \\
\midrule
G1 & Real-gaze & \cmark & -- & -- & 85.25 & 93.00 & 84.30 & 88.40 & \abgain{+0.91 / +1.50} \\
G2 & Real-gaze & \cmark & \cmark & -- & 85.78 & 93.80 & 84.20 & 88.74 & \abgain{+1.44 / +1.84} \\
G3 & Real-gaze & \cmark & -- & \cmark & 85.92 & 94.10 & 84.20 & 88.87 & \abgain{+1.58 / +1.97} \\
\rowcolor{gray!15}
G4 & Real-gaze & \cmark & \cmark & \cmark & 86.18 & 94.64 & 84.00 & 89.00 & \abgain{+1.84 / +2.10} \\
\midrule
M1 & Box-derived & \cmark & -- & -- & 84.95 & 91.80 & 84.30 & 87.89 & \abgain{+0.61 / +0.99} \\
M2 & Box-derived & \cmark & \cmark & -- & 85.48 & 93.00 & 84.10 & 88.33 & \abgain{+1.14 / +1.43} \\
M3 & Box-derived & \cmark & -- & \cmark & 85.57 & 93.20 & 84.20 & 88.47 & \abgain{+1.23 / +1.57} \\
\rowcolor{gray!10}
M4 & Box-derived & \cmark & \cmark & \cmark & 85.76 & 94.12 & 84.00 & 88.77 & \abgain{+1.42 / +1.87} \\
\bottomrule
\end{tabular}
\end{adjustbox}
\vspace{-4mm}
\end{table*}

Instead, it provides a relative spatial ranking of plausible candidate regions.
RPFM uses this ranking to modulate multi-scale features, while PAQI converts its local high-priority peaks into additional anchor queries for decoder localization.
Therefore, the visualization supports the intended function of the learned priority representation: it provides image-conditioned pre-localization guidance without requiring real or pseudo-gaze input during inference.
Because the displayed examples are selected qualitative cases, they complement rather than replace the quantitative detection and ablation results.

\subsection{Generalization to Other Detector Families}
\label{sec:detector_family_generalization}

\begin{table*}[t]
\centering
\setlength{\tabcolsep}{3.0pt}
\renewcommand{\arraystretch}{1.08}
\caption{Detector-family generalization on TIR-UAV120-Gaze and Anti-UAV410. Priority-supervised variants are compared with their reproduced no-priority baselines. Green and red values denote absolute gains and decreases within the same dataset and detector family, respectively, and bold values denote the best result within each detector-family block.}
\label{tab:detector_family_generalization_ablation_expanded}
\begin{adjustbox}{max width=\textwidth}
\begin{tabular}{ll l cc ccc c}
\toprule
\textbf{Dataset} &
\textbf{Base detector} &
\textbf{Priority-supervision map} &
\textbf{mAP$_{50}$} &
\textbf{$\Delta$ mAP$_{50}$} &
\textbf{P} &
\textbf{R} &
\textbf{F1} &
\textbf{$\Delta$ F1} \\
\midrule

\multirow{9}{*}{\makecell[c]{TIR-UAV120-\\Gaze}}
& \multirow{3}{*}{DINO-DETR}
& None & 84.34 & -- & 88.19 & 85.65 & 86.90 & -- \\
& & Box-derived & 85.76 & \abgain{+1.42} & 94.12 & 84.00 & 88.77 & \abgain{+1.87} \\
& & Real-gaze & \textbf{86.18} & \abgain{+1.84} & 94.64 & 84.00 & \textbf{89.00} & \abgain{+2.10} \\
\cmidrule(lr){2-9}
& \multirow{3}{*}{RT-DETR-l}
& None & 79.34 & -- & 89.75 & 80.20 & 84.70 & -- \\
& & Box-derived & 81.92 & \abgain{+2.58} & 92.52 & 78.99 & \textbf{85.22} & \abgain{+0.52} \\
& & Real-gaze & \textbf{82.33} & \abgain{+2.99} & 92.27 & 79.16 & \textbf{85.22} & \abgain{+0.52} \\
\cmidrule(lr){2-9}
& \multirow{3}{*}{Deformable-DETR}
& None & 80.47 & -- & 96.24 & 81.59 & 88.31 & -- \\
& & Box-derived & \textbf{84.69} & \abgain{+4.22} & 95.41 & 83.00 & 88.77 & \abgain{+0.46} \\
& & Real-gaze & 84.39 & \abgain{+3.92} & 95.76 & 83.00 & \textbf{88.92} & \abgain{+0.61} \\
\midrule

\multirow{9}{*}{Anti-UAV410}
& \multirow{3}{*}{DINO-DETR}
& None & 83.37 & -- & 93.72 & 79.00 & 85.73 & -- \\
& & Box-derived & 87.06 & \abgain{+3.69} & 96.38 & 86.00 & \textbf{90.90} & \abgain{+5.17} \\
& & \makecell[l]{Transferred\\pseudo-gaze} & \textbf{87.08} & \abgain{+3.71} & 98.32 & 83.71 & 90.43 & \abgain{+4.70} \\
\cmidrule(lr){2-9}
& \multirow{3}{*}{RT-DETR-l}
& None & 86.42 & -- & 91.35 & 83.60 & 87.31 & -- \\
& & Box-derived & \textbf{87.80} & \abgain{+1.38} & 97.51 & 85.28 & \textbf{90.99} & \abgain{+3.68} \\
& & \makecell[l]{Transferred\\pseudo-gaze} & 87.63 & \abgain{+1.21} & 96.51 & 81.54 & 88.40 & \abgain{+1.09} \\
\cmidrule(lr){2-9}
& \multirow{3}{*}{Deformable-DETR}
& None & 81.43 & -- & 93.80 & 82.32 & 87.69 & -- \\
& & Box-derived & 84.29 & \abgain{+2.86} & 94.96 & 82.00 & \textbf{88.00} & \abgain{+0.31} \\
& & \makecell[l]{Transferred\\pseudo-gaze} & \textbf{84.34} & \abgain{+2.91} & 95.48 & 81.00 & 87.65 & \abdrop{0.04} \\
\bottomrule
\end{tabular}
\end{adjustbox}
\vspace{-2mm}
\end{table*}

Table~\ref{tab:detector_family_generalization_ablation_expanded} applies the same priority-supervision formulation to DINO-DETR, RT-DETR-l, and Deformable-DETR.
On TIR-UAV120-Gaze, both box-derived and real-gaze supervision improve the corresponding no-priority baselines for all three detector families.
The mAP$_{50}$ gains range from 1.42 to 4.22 points, while the F1 gains range from 0.46 to 2.10 points.
Real-gaze supervision gives the highest mAP$_{50}$ for DINO-DETR and RT-DETR-l, whereas box-derived supervision is slightly better for Deformable-DETR.
Thus, the relative behavior of the two supervision schemes depends on the detector architecture and evaluation metric.

On Anti-UAV410, both box-derived and transferred pseudo-gaze supervision improve mAP$_{50}$ over the corresponding no-priority baselines across all three detector families.
Box-derived supervision also improves F1 consistently, whereas the transferred pseudo-gaze variant produces detector-dependent F1 changes, including a negligible 0.04-point decrease with Deformable-DETR.
Transferred pseudo-gaze supervision nevertheless gives slightly higher mAP$_{50}$ for DINO-DETR and Deformable-DETR.
Taken together, these results show that explicit priority supervision generalizes beyond DINO-DETR, while the preferred supervision-map source remains detector- and metric-dependent.

\subsection{Discussion and Limitations}
\label{sec:discussion_limitations}

The experiments consistently support the main claim that an explicitly supervised spatial priority representation can complement conventional classification and localization losses.
Across two datasets and three detector families, the clearest improvements occur in mAP$_{50}$, precision, and F1.
Meanwhile, the module ablations show that the auxiliary priority objective, feature modulation, and query injection make distinct contributions.
The attribute and visualization analyses further show that the predicted maps can retain target-related peaks in complex scenes, although their spatial distributions may include secondary responses around plausible distractors.

The results do not establish a universal ordering between box-derived, real-gaze, and transferred pseudo-gaze supervision.
Real gaze performs best on the main TIR-UAV120-Gaze comparison, but box-derived supervision gives the highest F1 on Anti-UAV410 and is preferable for some detector--metric combinations.
Accordingly, gaze should be viewed as an additional source of task-driven spatial information rather than as a guaranteed replacement for low-cost box-derived priority-supervision maps.

Pseudo-gaze transfer also has a clear practical boundary.
It avoids collecting new eye-tracking data, but its quality depends on the gaze--box statistics learned from TIR-UAV120-Gaze and on the domain gap to Anti-UAV410.
The more diffuse maps in several Anti-UAV410 examples are consistent with this limitation.
In addition, PAQI currently uses fixed weak-target size priors and non-differentiable peak selection, which may be less suitable when the predicted priority distribution is diffuse or the test target scale differs from the training priors.

Finally, the normalized predicted priority map represents relative spatial ranking rather than target-presence probability.
On a target-absent frame, the detection classifier must still assign priority-guided queries to the no-object class.
Future work can therefore investigate an explicit no-target state, adaptive size priors, differentiable priority-to-query conversion, and temporal priority modeling while preserving image-only inference.

\section{Conclusion}
\label{sec:conclusion}

This paper addresses thermal infrared weak-small UAV detection from the perspective of the gap between final localization labels and the search process preceding localization.
To provide explicit pre-localization guidance, we propose Gaze-DETR, which predicts an internal priority map from infrared image features.
RPFM uses this map for residual multi-scale feature modulation, whereas PAQI converts its high-priority locations into additional anchor queries for decoder localization.
Together, these components connect priority prediction with both feature processing and candidate localization while retaining image-only inference.

To train the priority head under different annotation conditions, we introduce three priority-supervision schemes: box-derived Gaussian supervision, real-gaze fixation-density supervision, and transferred pseudo-gaze supervision based on statistical gaze--box relation modeling.
To support the latter two schemes, we construct TIR-UAV120-Gaze with paired thermal infrared detection and task-driven eye-tracking annotations.
The transferred relation is further applied to the Anti-UAV410 training boxes when real eye-tracking annotations are unavailable.
Across all settings, priority-supervision maps are used only during training, and the network architecture remains unchanged.

Experiments on TIR-UAV120-Gaze and Anti-UAV410 show that the evaluated priority-supervision settings improve the corresponding DINO-DETR baselines.
The module, attribute, visualization, and cross-detector analyses further support the role of explicit spatial-priority learning, while also showing that the relative behavior of different priority-supervision-map sources depends on the dataset, detector, and evaluation metric.
Taken together, these results support a conservative conclusion: priority supervision provides effective pre-localization guidance that complements conventional bounding-box localization supervision under different annotation conditions and costs.

\FloatBarrier
\bibliographystyle{IEEEtran}
\bibliography{sn-bibliography}

\end{document}